\documentclass[accepted,specialissue]{melba}

\usepackage{mwe} 

\usepackage{amsmath,amsfonts}

\usepackage{cite}
\usepackage{amsmath,amssymb,amsfonts}
\usepackage{algorithmic}
\usepackage{graphicx}
\usepackage{booktabs}
\usepackage{textcomp}
\usepackage{multirow}
\usepackage{multicol}
\usepackage{xcolor}
\usepackage{hyperref}
\usepackage{caption}
\usepackage{float}
\usepackage{placeins}
\usepackage{rotating}
\usepackage{subcaption}

\melbaid{2025:025}  
\doi{10.59275/j.melba.2025-1e7f}
\melbaauthors{Gaur, Vitale, Hering, Kwisthout, Jacobs, Philipp, and van der Graaf}  
\email{fennie.vandergraaf@radboudumc.nl}
\volume{3}
\firstpageno{559}  
\melbayear{2025}  
\datesubmitted{2025-04-18}  
\datepublished{2025-11}  

\melbaspecialissue{Special issue on Fairness of AI in Medical Imaging (FAIMI)}
\melbaspecialissueeditors{Veronika Cheplygina, Aasa Feragen, Andrew King, Ben Glocker, Enzo Ferrante, Eike Petersen, Esther Puyol-Antón, Melanie Ganz-Benjaminsen}

\ShortHeadings{Fairness Evaluation of Risk Estimation Models for Lung Cancer Screening}{Gaur, Vitale, Hering, Kwisthout, Jacobs, Philipp, and van der Graaf}

\title{Fairness Evaluation of Risk Estimation Models \\ for Lung Cancer Screening}

\author{
	\firstname Shaurya \surname Gaur\aff{1,2}\orcid{0009-0004-7740-959X},
	\firstname Michel \surname Vitale\aff{1,3}\orcid{0009-0000-5476-7504},
    \firstname Alessa \surname Hering\aff{1}\orcid{0000-0002-7602-803X},
    \firstname Johan \surname Kwisthout\aff{2}\orcid{0000-0003-4383-7786},
	\firstname Colin \surname Jacobs\aff{1}\orcid{0000-0003-1180-3805},
    \firstname Lena \surname Philipp\aff{1}\orcid{0009-0005-2742-7041}$^{*}$,
    \firstname Fennie \surname van der Graaf\aff{1}\orcid{0000-0002-9781-4101}$^{*}$,
}
\affiliations{
	\num 1 \addr Diagnostic Image Analysis Group, Department of Medical Imaging, Radboud University Medical Center, \\ Nijmegen, the Netherlands \\
    \num 2 \addr Department of Artificial Intelligence, Radboud University, Nijmegen, the Netherlands \\
    \num 3 \addr Ethics of Healthcare Group, Department IQ Health, Radboud University Medical Center, Nijmegen, the Netherlands \\
}

\abstract{
	Lung cancer is the leading cause of cancer-related mortality in adults worldwide. Screening high-risk individuals with annual low-dose CT (LDCT) can support earlier detection and reduce deaths, but widespread implementation may strain the already limited radiology workforce. Artificial intelligence (AI) models have shown potential in estimating lung cancer risk from LDCT scans. However, high-risk populations for lung cancer are diverse, and these models' performance across diverse demographic groups remains an open question.  
\revision{In this study, we drew on the considerations on confounding factors and ethically significant biases outlined in the JustEFAB framework to evaluate} potential performance disparities and fairness in two \revision{deep learning} risk estimation models for lung cancer screening: the Sybil lung cancer risk model and the Venkadesh21 nodule risk estimator. We also examined disparities in the PanCan2b logistic regression model recommended in the British Thoracic Society nodule management guideline. Both \revision{deep learning} models were trained on data from the U.S.-based National Lung Screening Trial (NLST), and assessed on a held-out NLST validation set. We evaluated area under the ROC curve (AUROC), sensitivity, and specificity across demographic subgroups, and explored potential confounding from clinical risk factors.
We observed a statistically significant AUROC difference in Sybil’s performance between women (0.88, 95\% CI: 0.86, 0.90) and men (0.81, 95\% CI: 0.78, 0.84, $p < .001$). At 90\% specificity, Venkadesh21 showed lower sensitivity for Black (0.39, 95\% CI: 0.23, 0.59) than White participants (0.69, 95\% CI: 0.65, 0.73). These differences were not explained by available clinical confounders and \revision{thus} may be classified as unfair biases according to JustEFAB. Our findings highlight the importance of improving and monitoring model performance across underrepresented subgroups in lung cancer screening, as well as further research on algorithmic fairness in this field.}


\begin{document}

\twocolumn[\maketitle]

\begingroup
\renewcommand\thefootnote{*}
\footnotetext{Contributed equally as last authors.}
\endgroup


\section{Introduction}

Lung cancer is the leading cause of cancer-related death in adults worldwide. To treat lung cancer before it becomes terminal \citep{Newson_2011}, radiologists have explored the use of screenings on high-risk populations, collecting images of their lungs to assess cancer risk by detecting pulmonary nodules and assessing their risk of malignancy \citep{dlcst}. Results from the U.S. National Lung Screening Trial (NLST) \citep{nlst} and the Dutch-Belgian NELSON trial \citep{deKoning2020-NELSON} found that lung cancer screening (LCS) with low-dose computed tomography (LDCT) scans can significantly reduce lung cancer mortality in a high-risk population. Findings like these have encouraged trials and adoptions of LCS programs around the world at both the regional and national level\footnote{\url{https://www.lungcancerpolicynetwork.com/interactive-map-of-lung-cancer-screening/}}. However, considering a global shortage of radiologists \citep{Konstantinidis2024-RadShortage}, implementing LCS would exacerbate their already stressful workload \citep{Bruls2020-RadWorkload}, increasing the risk of burnout and potentially reducing accuracy \citep{Harolds2016-RadBurnout}. 

To reduce radiologist workloads, several artificial intelligence (AI) systems have been built to automatically detect \citep{Hendrix2023-NoduleDetection}, measure and classify nodules \citep{Lancaster2025-AIsystem}, and assess the risk of malignancy of nodules \citep{venkadesh_deep_2021} and of lung cancer in LDCT scans \citep{mikhael_sybil_2023}. These deep learning (DL) models are trained and evaluated on LDCT datasets from screening trials \citep{nlst, dlcst} and have been shown to perform comparably to radiologists in nodule malignancy risk estimation \citep{venkadesh_deep_2021}. Some models have even shown greater performance than the Brock model \citep{venkadesh_deep_2021}, a nodule malignancy calculator based on various lung cancer risk factors which has been employed by radiologists in screening protocols \citep{Lim2020-ilst, Baldwin2016-BTS}.

However, AI systems have shown biases between demographic subgroups based on factors like race, sex, and socioeconomic status \citep{obemeyer2019riskcostproxy, seyyed-kalantari_underdiagnosis_2021}, which could perpetuate historical inequalities in healthcare \citep{Spencer2013-biasInsurance}. In particular, DL models have tended to show lower accuracy for groups with less representation in their development data, such as women and racial minority groups \citep{seyyed-kalantari_underdiagnosis_2021}. These groups have also been underrepresented in LCS trials \citep{nlst, dlcst} and sometimes even excluded from analysis \citep{deKoning2020-NELSON}, often due to biased screening criteria. LCS protocols, like the 2013 US Preventative Services Task Force guidelines, chose who to screen based on risk factors like age (i.e. 55 $-$ 80 years old) and smoking history (i.e. $\geqslant$ 30 pack-years of smoking) \citep{Moyer2014-USPSTF2013}. Studies have shown that these guidelines are biased against women and Black Americans, who are less likely to clear the 30 pack-year threshold \citep{Shum2024-GenderDisparityLCS, Aredo2022-RaceDisparityLCS}. This is especially concerning since studies have observed rising lung cancer rates for women even with lower smoking rates \citep{siegfriend-gender-lc-copd} and a higher lung cancer risk for Black American men than men in other racial groups \citep{deGroot2018-EducationSmoking}.

Bias in model performance between groups can also result from decisions made during model development, such as the choice of input data, labels, network architecture, and the loss functions upon which to optimize a model \citep{ferrara2023Fairness, narayanan2018translation}. A model architecture may exhibit shortcut learning, or the ability to predict a demographic characteristic in a way that may lead to spurious correlations which could bias performance for certain subgroups \citep{Brown2023-ShortcutLearning}. For example, \citet{mikhael_sybil_2023} demonstrated that the Sybil lung cancer risk estimation model was able to accurately classify a participant's sex, height, weight, and smoking status from an LDCT. Additionally, clinical decisions made based on model output, such as the threshold used to stratify risk scores as either benign or malignant, can lead to disparate impact between subgroups \citep{ferrara2023Fairness}. In LCS, such biases could lead to the overestimation or underestimation of lung cancer risk. Overestimation can unnecessarily burden both participants and healthcare practitioners with further treatment, with the potential to result in overdiagnosis for a particular group. Conversely, underestimation can falsely flag people with lung cancer as low-risk, potentially barring them from life-saving treatment, increasing the risk of underdiagnosis for a community. These biases would not only materially harm the affected communities based on their demographics, but also reduce their trust in LCS \citep{Newson_2011}.

Usually, these biases are considered to be \textit{unfair} when they lead an AI model to systematic (and unwarranted) differences in behavior towards a specific demographic subgroup of participants. While bias can be observed statistically, algorithmic fairness depends on stakeholders' values and priorities \citep{narayanan2018translation}. This has led to various fairness definitions and frameworks \citep{obermeyer2021algorithmic, mccraden_justefab}, which prioritize equalizing different metrics between groups. For example, the Algorithmic Bias Playbook from \citet{obermeyer2021algorithmic} prioritizes examining the differences in probability calibration curves between subgroups and choosing fair labels for the prediction task. Alternatively, the JustEFAB framework from \citet{mccraden_justefab} emphasizes defining biases based on metrics which would measure treatment outcomes. Instead of examining fairness between groups, some studies argue that fairness should be examined more on an individual level, and ensure similar cases are treated alike \citep{giovanola_beyond_2023}. Multiple studies have demonstrated that metrics between different AI fairness frameworks can be conflicting, so a framework should be carefully chosen according to its use case \citep{barocas-hardt-narayanan, Chouldechova2020-impossibility, richardson2021framework}. 

Moreover, a fairness evaluation in healthcare must properly consider any links between demographic groups and clinical confounders. \citet{Jones2024} describes how medical imaging datasets often have demographic disparities in the prevalence of clinical risk factors, the presentation of medical images, and the annotation practices surrounding malignancy labels. These clinical disparities add a layer of dataset bias that may confound a performance disparity seen between subgroups \citep{Bernhardt2022-UnderdiagnosisBiasResponse}. To assess fairness based on these links, we must carefully examine their causality: if a demographic is clinically linked to a medical risk factor, it may be a fair bias, and this trait may be a useful predictor. This reflects how clinicians aim to use all of the information about a patient (including demographic traits) to gain valuable insights about their risk for disease \citep{giovanola_beyond_2023}. Conversely, some prevalence links may be due to unfair selection biases, such as racial disparities in access to healthcare, which influence who in a particular group is more likely to have a medical image taken \citep{Jones2024}.

\revision{In this study, we were guided by the first part of the JustEFAB framework and its considerations on performance analysis and ethically significant biases. We used this part to evaluate the fairness of two DL models for LCS: a pulmonary nodule risk estimation model from} \citet{venkadesh_deep_2021}, and Sybil, a lung cancer risk estimation model from \citet{mikhael_sybil_2023}. We also evaluate the PanCan2b model from \citet{mcwilliams_probability_2013}, a multivariate logistic regressor popular in lung cancer screening protocols. We first examine whether our models have performance disparities between different groups in our datasets based on demographic characteristics. If any such biases have been found, we then assess prevalence disparities for clinical risk factors between subgroups, and examine performance when isolating for these confounders. We assess these biases according to the JustEFAB ethical framework for integrating machine learning in clinical practice \citep{mccraden_justefab}, and briefly reflect on future research needed in the field of algorithmic fairness.
\section{Methods}


This study aims to evaluate the fairness of three risk estimation models for lung cancer screening on the NLST cohort. We performed a two-stage analysis, first investigating disparities in model performance between demographic subgroups, and then evaluating the fairness of these disparities based on confounding from lung cancer risk factors. The code for this analysis is publicly accessible online\footnote{\url{https://github.com/DIAGNijmegen/bodyct-lung-malignancy-fairness}}.

\subsection{NLST Dataset} \label{sec:datasets}

This retrospective study evaluated fairness using LDCT examinations from NLST participants \citep{nlst}. NLST was a multi-center randomized controlled clinical trial which screened participants \revision{at 33 different centres in the USA} between 2002 and 2007. In total, 53,454 participants were recruited, and 26,722 participants were randomly assigned to receive three annual screenings with low-dose chest CT. These participants were selected as high-risk on account of being between the ages of 55 and 74 years old and having smoked at least 30 pack-years of cigarettes by the start of the trial. NLST participants completed a wide-ranging questionnaire before screening, reporting demographic characteristics, smoking behavior, histories of work, diseases disease and previous cancers, and their family's history of lung cancer. Information regarding the status, type, diagnosis, and survival outcomes of lung cancer was also recorded during screening. This trial received approval from the institutional review boards from all 33 centers involved, and obtained informed consent from all participants. Permission to use the data for this study was obtained from the NLST through the National Cancer Institute Cancer Data Access System (approved Project ID: NLST-1268).

\begin{table*}[ht!]
\centering
\caption{Demographic characteristics of the NLST validation set (n=5911 scans). HS = High School.}
\label{tab:datasetDemos}
\begin{tabular}{ll|rrr}
\toprule
Characteristic & Subgroup & Malignant (n=581) & Benign (n=5330) & All Scans (n=5911) \\
\midrule
\multirow[c]{7}{*}{Education Status} & 8th grade or less & 9 (1.5) & 102 (1.9) & 111 (1.9) \\
 & 9th-11th grade & 32 (5.5) & 258 (4.8) & 290 (4.9) \\
 & Associate Degree & 126 (21.7) & 1175 (22.0) & 1301 (22.0) \\
 & Bachelors Degree & 96 (16.5) & 817 (15.3) & 913 (15.4) \\
 & Graduate School & 76 (13.1) & 778 (14.6) & 854 (14.4) \\
 & HS Graduate / GED & 141 (24.3) & 1338 (25.1) & 1479 (25.0) \\
 & Post-HS training & 87 (15.0) & 765 (14.4) & 852 (14.4) \\
\cline{1-5}
\multirow[c]{6}{*}{Race} & Asian & 6 (1.0) & 66 (1.2) & 72 (1.2) \\
 & Black & 28 (4.8) & 160 (3.0) & 188 (3.2) \\
 & More than one race & 6 (1.0) & 59 (1.1) & 65 (1.1) \\
 & Native American & 8 (1.4) & 15 (0.3) & 23 (0.4) \\
 & Native Hawaiian & 1 (0.2) & 17 (0.3) & 18 (0.3) \\
 & White & 530 (91.2) & 4993 (93.7) & 5523 (93.4) \\
\cline{1-5}
\multirow[c]{2}{*}{Ethnicity} & Hispanic/Latino & 4 (0.7) & 94 (1.8) & 98 (1.7) \\
 & Not Hispanic/Latino & 574 (98.8) & 5205 (97.7) & 5779 (97.8) \\
\cline{1-5}
\multirow[c]{2}{*}{Sex} & Female & 244 (42.0) & 2226 (41.8) & 2470 (41.8) \\
 & Male & 337 (58.0) & 3104 (58.2) & 3441 (58.2) \\
\cline{1-5}
Weight & Median (IQR) & 175 (50) & 180 (50) & 180 (50) \\
\cline{1-5}
Height & Median (IQR) & 68 (6) & 68 (6) & 68 (6) \\
\cline{1-5}
Body Mass Index & Median (IQR) & 26 (4) & 27 (6) & 26 (6) \\
\cline{1-5}
Age & Median (IQR) & 64 (8) & 62 (8) & 62 (8) \\
\cline{1-5}
\bottomrule
\end{tabular}
\end{table*}

Both DL models were trained on subsets of NLST data \citep{venkadesh_deep_2021, mikhael_sybil_2023}. While these models were trained on different NLST subsets, their training sets contain very similar demographic characteristics, shown in Tables \ref{tab:venk21trainData} and \ref{tab:sybilTrainData} in the Appendix.
\revision{These models were chosen because they were both trained using NLST data, this motivated us to focus our investigation on whether the cause of biases was likely related to disparities in the dataset (if models shared the same bias), or whether they likely stemmed from differences in the model training process (if one model exhibited a bias not observed in another model).}
We performed an internal validation of our models' biases using a subset of NLST data at the scan level, since Sybil does not return nodule-level risk estimates. For the Venkadesh21 model and PanCan2b, we employed the maximum malignancy score for a nodule as its score for a LDCT scan. Our internal validation subset included all scans from the dataset from \citet{venkadesh_deep_2021}, excluding any scans found in Sybil's training set. This, along with using purely predictions from validation folds from the Venkadesh21 model, ensured that no predictions are from the training data itself. After this selection, our validation set consisted of 5,911 scans from 3,492 participants to evaluate the Venkadesh21 model, Sybil, and PanCan2b. Demographic characteristics of this scan-level dataset are included in Table \ref{tab:datasetDemos}.

\subsection{Risk Estimation Models} \label{sec:models}


We analyzed two deep learning risk estimation models in this study both trained on NLST \citep{nlst}, along with the PanCan2b risk calculator trained on the PanCan cohort. We used the models as specified in their original publications \citep{venkadesh_deep_2021, mikhael_sybil_2023, mcwilliams_probability_2013} and in relevant supplementary material, to prioritize understanding each model's potential biases in their publicly-available state. All models return risk scores between $0.0$ and $1.0$. An overview of each model's training dataset size, input type and output is in Table \ref{tab:modelBasics}. 

\begin{table*}[htb!]
    \centering
    \caption{Overview of inputs and outputs for the models. $N$ refers to the amount of data of the specified input type.}
    \begin{tabular}{ccrrl}
    \toprule
        Model       & Input Type                                & \multicolumn{2}{c}{$N$ (Train / Test)} & Additional Information \\
    \midrule
        Venkadesh21 & $50mm^3$ Block Around Nodule in LDCT      & 16,077 &           $-$  & Trained with 10-fold cross-val. \\
        Sybil       & Full LDCT (+ training annotations)        & 26,182 &           13,121 & Outputs scores for next 6 years. \\
        PanCan2b    & Nodule, Scan and Participant Characteristics  &  7,008 &            5,021 & Trained on the PanCan trial. \\
    \bottomrule
    \end{tabular}
    \label{tab:modelBasics}
\end{table*}

\subsubsection{Venkadesh21}

The first model we analyzed in this study was a malignancy risk estimation model for pulmonary nodules developed by \citet{venkadesh_deep_2021}. This model takes a LDCT scan and XYZ-coordinates of previously detected pulmonary nodules, annotated by an expert radiologist and two trained medical students. It then extracts a $50mm^3$ patch around the nodules and this 3D patch is subsequently processed by two deep learning models. The first model extracts features from nine 2D views of the nodule area using a ResNet50 CNN backbone and combines them with a fully-connected layer to extract a malignancy risk score. The second model extracts features from the 3D nodule volume using a 3D Inception-v1 CNN and processes them using a linear layer to obtain a single risk score. The Venkadesh21 model was trained using 10-fold cross validation and the final score consisted of an average of the 20 resulting models. It was optimized using a cross entropy loss, comparing predictions to data from NLST regarding whether lung cancer was found. The final outputs are calibrated\footnote{\label{venk21cal}\url{https://www.diagnijmegen.nl/software/nodule-malignancy-risk-calibration/}} using a two-parameter logistic regression with Platt scaling.

This DL model demonstrated comparable performance to thoracic radiologists, and significantly outperformed PanCan2b, when examining area under the receiver operator characteristic curve (AUROC) when evaluated on an independent cohort of nodules from the Danish Lung Cancer Screening Trial \citep{dlcst}. This algorithm is freely available online for research purposes\footnote{\url{https://grand-challenge.org/algorithms/pulmonary-nodule-malignancy-prediction/}}.

\subsubsection{Sybil}

Sybil is a deep learning model for estimating lung cancer risk for the next six years after a given full LDCT scan, developed by \citet{mikhael_sybil_2023}. An LDCT image is passed through a ResNet3D encoder to obtain embeddings which are then passed through two separate layers. The first is a 3D max pooling layer. The second is an attention-based pooling layer that identifies features from CT slices and regions most relevant to Sybil's lung cancer risk predictions. The outputs of both of these layers are concatenated and ran through a hazard layer, which calculates cumulative probabilities of lung cancer risk over the six years after a LDCT. Sybil was trained five times with final predictions as the average of this ensemble, with three components to its loss function. The first is a cross entropy loss comparing Sybil's predictions for each of the next six years with whether a diagnosis happened at that time, according to NLST. The other two components correspond to whether the attention aligned with any included information about the cancerous lung and the bounding box annotations. Finally, the outputs are calibrated using an isotonic regressor. Sybil performed well when evaluated on a NLST test set and screening datasets from two hospitals in the USA and Taiwan \citep{mikhael_sybil_2023}, and the model and code are available freely online\footnote{\url{https://github.com/reginabarzilaygroup/Sybil.git}}. We used the Year 1 scores from Sybil in our analysis, since this score was the most directly comparable to estimations of the other models, which aim to predict the current risk of lung cancer in an LDCT at baseline. 

\subsubsection{PanCan2b \revision{(Brock Model)}}

We also assessed the fairness of the \revision{Brock} nodule malignancy risk estimation model developed by \citet{mcwilliams_probability_2013}\revision{, in particular its PanCan2b variation}. This is a multivariate logistic regression model based on statistical analysis of lung cancer risk factors in a dataset from the Pan-Canadian Early Detection of Lung Cancer study (PanCan), containing 7,008 nodules from 2,537 participants. Factors included in this analysis are those available in the dataset which were known at the time to be associated with lung cancer, along with nodule characteristics that can be determined from an LDCT at the time of screening. From this analysis, \citet{mcwilliams_probability_2013} developed a parsimonious model "PanCan1a" which significantly links female sex, larger nodules, and nodules in the upper lung with lung cancer risk ($p < 0.05$). They also developed a full model "PanCan2a" which added factors significant at $p < 0.25$: age, family history of LC, emphysema, nodule type, and the number of nodules per scan. Additional models including nodule spiculation were also developed ("PanCan1b" and "PanCan2b").

These models' performances were validated on participants from a trial in British Columbia with AUROC scores above 0.90. They have subsequently been validated in various settings and have shown strong overall performance, with the full model achieving higher AUROC than the parsimonious model \citep{Chen2025-BrockMetaAnalysis}. The Brock model has been applied in the nodule management guideline for the International Lung Screening Trial (ILST) \citep{Lim2020-ilst}, with the PanCan2b full model with spiculation recommended by the British Thoracic Society for nodule management \citep{Baldwin2016-BTS}. In the ILST protocol, a PanCan2b malignancy risk score of 6\% (0.06) or above is associated with a moderate malignancy risk, and a follow-up LDCT is recommended after 3 months, instead of an annual or biennial LDCT at lower risk levels \citep{Lim2020-ilst}.

\subsection{JustEFAB Ethical Framework}

For this study, we applied \revision{part of} the \textit{Justice, Equity, Fairness, and Anti-Bias} (JustEFAB) group fairness framework as proposed by \citet{mccraden_justefab}, which recognizes the sociotechnical character of AI systems and draws on principles of medical and research ethics, feminist philosophy of science, and theories of justice.
Like many other ethical frameworks in AI fairness \citep{obermeyer2021algorithmic, richardson2021framework, giovanola_beyond_2023}, JustEFAB examines the entire lifecycle of AI models, though we believe its careful ethical consideration of the links between demographics and confounders is particularly relevant for LCS. From JustEFAB, we primarily focused on Stage 1A (Design and Development) and Stage 1B (Ethical Decision Making), primarily analyzing the fairness of the data used by the models and choices made during training. We were less concerned with bias in the NLST label annotations, since evaluating the NLST protocol for diagnosing lung cancer was out of scope for this study, so we therefore assumed the provided labels truly reflect lung cancer.

For this study, we adopted the definition of algorithmic bias as written in JustEFAB, referring to a performance disparity leading to disparate impacts for a demographic group. We primarily defined fairness as a lack of \textit{ethically significant bias}, defined in JustEFAB as a performance disparity which would systematically have negative consequences for a demographic group's treatment based on their identity and not on clinical need \citep{mccraden_justefab}. \revision{Within this context, we use the separation criteria for AI fairness as defined by \citet{barocas-hardt-narayanan} within the JustEFAB framing, controlling for clinical need by examining potential confounders.} In this LCS context, we were primarily concerned with minimizing two negative outcomes: overestimation (false positives) and underestimation (false negatives) of lung cancer risk. 

\begin{figure*}[t]
    \centering
    \includegraphics[width=0.95\linewidth]{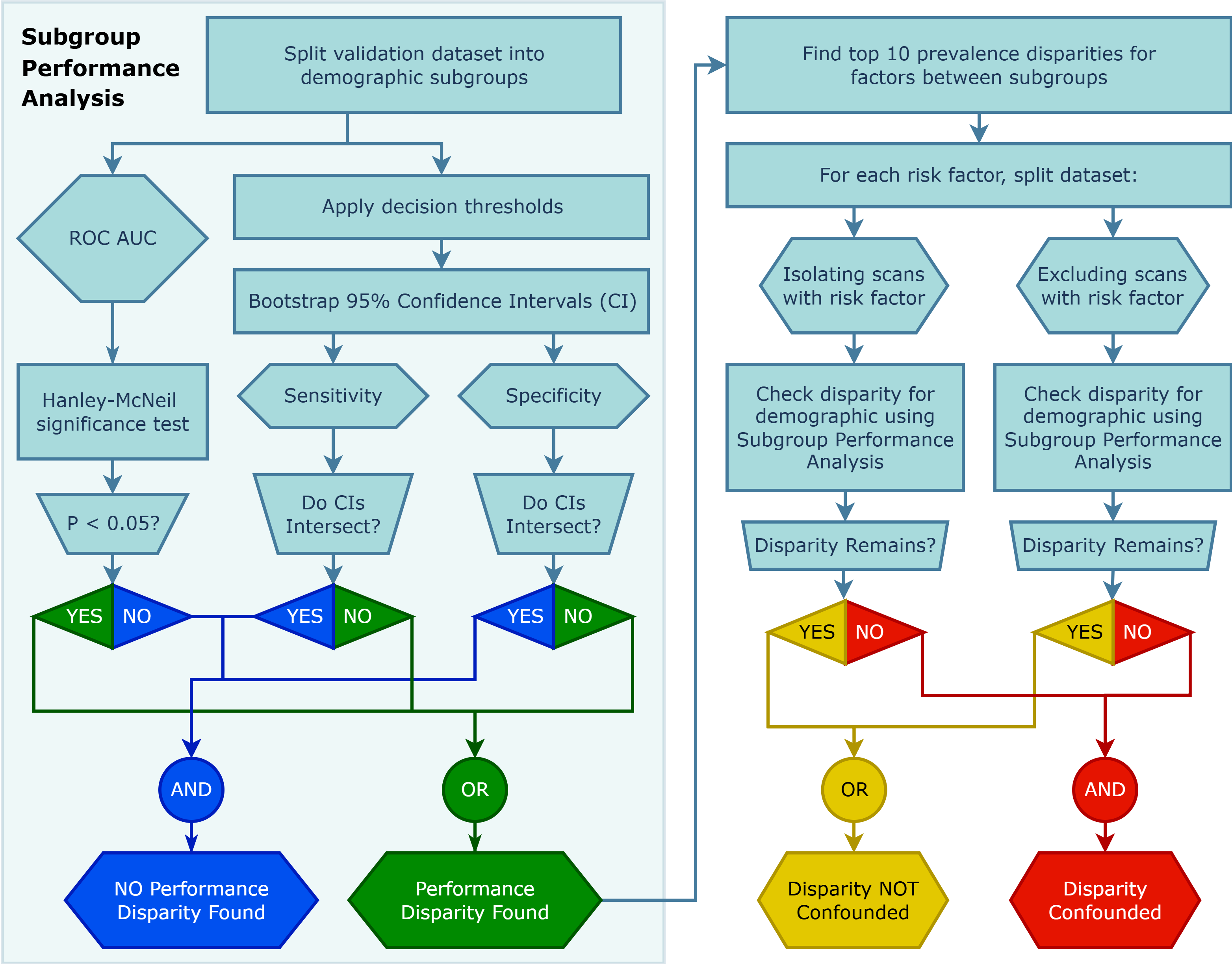}
    \caption{Flowchart of subgroup performance analysis (left) and fairness assessment (right) for a single model on a single demographic characteristic. Color coding indicates the decision paths to follow to reach a result (ex. disparity found).}
    \label{fig:flowchart}
\end{figure*}

\subsection{Stage 1: Subgroup Performance Analysis}


Based on demographic characteristics from NLST, we split the validation cohort into separate subgroups, and evaluated the models' performance on each subgroup. Figure \ref{fig:flowchart} displays the process for performing this analysis for one model on a single characteristic.
For binary categories (such as sex and race), we compared performance directly between groups, while with numerical characteristics (i.e. age, weight, height), we split the cohort based on the median in the overall NLST dataset (see Table \ref{tab:datasetDemos}). We calculated body mass index (BMI) using the provided height and weight, and split the cohort based on the 25 $kg/m^2$ threshold between normal-weight and overweight people, as recognized by the World Health Organization \citep{WHOsurfreport-BMI}. For education, we divided the cohort between those who received a high school diploma, GED, or higher, and those who had only received an education up to the 11th grade.

We evaluated overall performance by comparing \revision{the AUROC} for each model between subgroups. Significance was determined with a two-tailed hypothesis test ($\alpha = 0.05$) from \citet{HanleyMcNeil1982-ROC}. This test between two independent populations accounts for class imbalance by directly considering the number of malignant and benign scans from each group in its calculation. 

To examine underestimation and overestimation of lung cancer risk, we also compared model sensitivity and specificity between subgroups. This was conducted at three thresholds, set for each model based on the overall validation set: 90\% \revision{overall} sensitivity, 90\% \revision{overall} specificity, and the "moderate risk" threshold of 6\% used for the Brock model, \revision{which is being applied in} the ILST \citep{Lim2020-ilst}. Table \ref{tab:thresholds} displays the thresholds on our NLST validation cohort. We collected 95\% confidence intervals (CIs) for ROC curves, sensitivity and specificity using 1000 bootstraps. We evaluated a disparity in sensitivity and specificity as \textit{substantial} when a model's CIs between subgroups do not intersect.

\begin{table}[ht!]
\centering
\caption{Thresholds used to evaluate model performance on NLST. All models are also evaluated on the Brock ILST 6\% (0.06) moderate risk threshold \citep{Lim2020-ilst}.}
\label{tab:thresholds}
\begin{tabular}{l|rr}
\toprule
 & 90\% Sensitivity & 90\% Specificity \\
\midrule
Venkadesh21 & 0.049 & 0.222 \\
Sybil (Year 1) & 0.003 & 0.058 \\
PanCan2b & 0.015 & 0.165 \\
\bottomrule
\end{tabular}
\end{table}

\subsection{Stage 2: Fairness Assessment}

If performance disparities between demographic subgroups for any model were discovered in Stage 1, our next step was to assess whether they are unfair, according to JustEFAB \citep{mccraden_justefab}. This allowed us to analyze whether the disparities are the result of biases in the training data or the trained algorithm, or just due to inherent challenges in the task across different groups. We assessed fairness by examining disparities in the prevalence of potential clinical confounders between subgroups. These included nodule and LDCT characteristics, participants' smoking and work histories in fields relevant to lung cancer, and previous diagnosis of diseases and cancers. Between subgroups, we measured a prevalence disparity as the difference in the percentage of participants with a particular risk factor. Here, we also split numerical risk factors based on their median in the NLST dataset, except for the number of nodules in a LDCT (split into bins of 1 nodule and more than 1 nodule).

We evaluated performance disparities between subgroups for the 10 factors with the highest prevalence disparity between them. The process for each factor is detailed in Figure \ref{fig:flowchart}. We employed the same metrics as in Stage 1, with a particular focus on the metric(s) for which a bias was observed. For each risk factor, we examined the model's performance between demographic subgroups in two NLST subsets: 1) the scans in which the risk factor was present, and 2) the scans where this factor was \textit{not} present. If the performance disparity from Stage 1 persisted in either subset, then the bias was determined to not be confounded by that factor. This persistence was based first on the $p$ value between AUROC and the non-intersection of CIs for sensitivity and specificity, and second on the difference in the performance metrics between demographic subgroups. If the performance disparity was no longer significant and reduced for both subsets, then the factor may be a potential confounder. However, assessing fairness with JustEFAB requires a careful examination of clinical need: if a factor is linked to the demographic through selection bias and not biology \citep{Jones2024}, or not related to lung cancer risk, then it is deemed unfair \citep{mccraden_justefab}.
\section{Results}

We first report the results of the subgroup performance analysis, highlighting demographic performance disparities identified for the models. Comprehensive results are provided in Tables \ref{tab:resNLSTscanROC} (AUROC), \ref{tab:resNLSTfullTPR} (sensitivity) and \ref{tab:resNLSTfullTNR} (specificity). For each demographic group where a performance disparity was observed, we subsequently analyzed potential contributing risk factors. Specifically, we report the most prevalent clinical confounders and evaluate model performance when stratifying by or excluding these variables.

\subsection{Subgroup Performance Analysis}

\begin{table*}[ht!]
\centering
\caption{AUROC (with 95\% confidence intervals) for models for demographic subgroups on NLST.}
\label{tab:resNLSTscanROC}
\begin{tabular}{ll|ll|ll|ll}
\toprule
 &  & \multicolumn{2}{c}{Venkadesh21} & \multicolumn{2}{c}{Sybil (Year 1)} & \multicolumn{2}{c}{PanCan2b} \\
Attribute & Group & AUROC & p & AUROC & p & AUROC & p \\
\midrule
\multirow[c]{2}{*}{Age} & $>$ 61 & 0.88 (0.86, 0.90) &  & 0.84 (0.82, 0.86) &  & 0.76 (0.74, 0.79) &  \\
 & $\leqslant$ 61 & 0.90 (0.88, 0.92) & .14 & 0.85 (0.82, 0.88) & .67 & 0.81 (0.78, 0.83) & .05 \\
\cline{1-8}
\multirow[c]{2}{*}{BMI} & $\geqslant$ 25 & 0.90 (0.88, 0.91) &  & 0.86 (0.84, 0.88) &  & 0.81 (0.79, 0.83) &  \\
 & $<$ 25 & 0.88 (0.85, 0.90) & .32 & 0.81 (0.78, 0.85) & .05 & 0.73 (0.70, 0.77) & .001 \\
\cline{1-8}
\multirow[c]{2}{*}{Education} & $\geqslant$ HS & 0.89 (0.88, 0.91) &  & 0.85 (0.83, 0.86) &  & 0.78 (0.76, 0.80) &  \\
 & $<$ HS & 0.85 (0.78, 0.92) & .32 & 0.82 (0.75, 0.89) & .55 & 0.78 (0.71, 0.85) & .98 \\
\cline{1-8}
\multirow[c]{2}{*}{Height} & $\leqslant$ 68 & 0.89 (0.87, 0.91) &  & 0.87 (0.85, 0.89) &  & 0.79 (0.76, 0.81) &  \\
 & $>$ 68 & 0.88 (0.86, 0.91) & .67 & 0.80 (0.77, 0.83) & $<$ .001 & 0.78 (0.75, 0.80) & .71 \\
\cline{1-8}
\multirow[c]{2}{*}{Race} & White & 0.89 (0.88, 0.91) &  & 0.84 (0.83, 0.86) &  & 0.78 (0.76, 0.80) &  \\
 & Black & 0.82 (0.74, 0.89) & .14 & 0.83 (0.74, 0.90) & .77 & 0.75 (0.65, 0.83) & .54 \\
\cline{1-8}
\multirow[c]{2}{*}{Sex} & Male & 0.89 (0.87, 0.91) &  & 0.81 (0.78, 0.84) &  & 0.79 (0.76, 0.81) &  \\
 & Female & 0.89 (0.87, 0.91) & .92 & 0.88 (0.86, 0.90) & $<$ .001 & 0.78 (0.75, 0.81) & .66 \\
\cline{1-8}
\multirow[c]{2}{*}{Weight} & $\leqslant$ 180 & 0.88 (0.87, 0.90) &  & 0.84 (0.82, 0.87) &  & 0.77 (0.74, 0.79) &  \\
 & $>$ 180 & 0.89 (0.87, 0.91) & .65 & 0.84 (0.81, 0.87) & .82 & 0.80 (0.77, 0.82) & .13 \\
\cline{1-8}
\bottomrule
\end{tabular}
\end{table*}

\begin{figure}[ht!]
    \centering
    \includegraphics[width=0.95\linewidth]{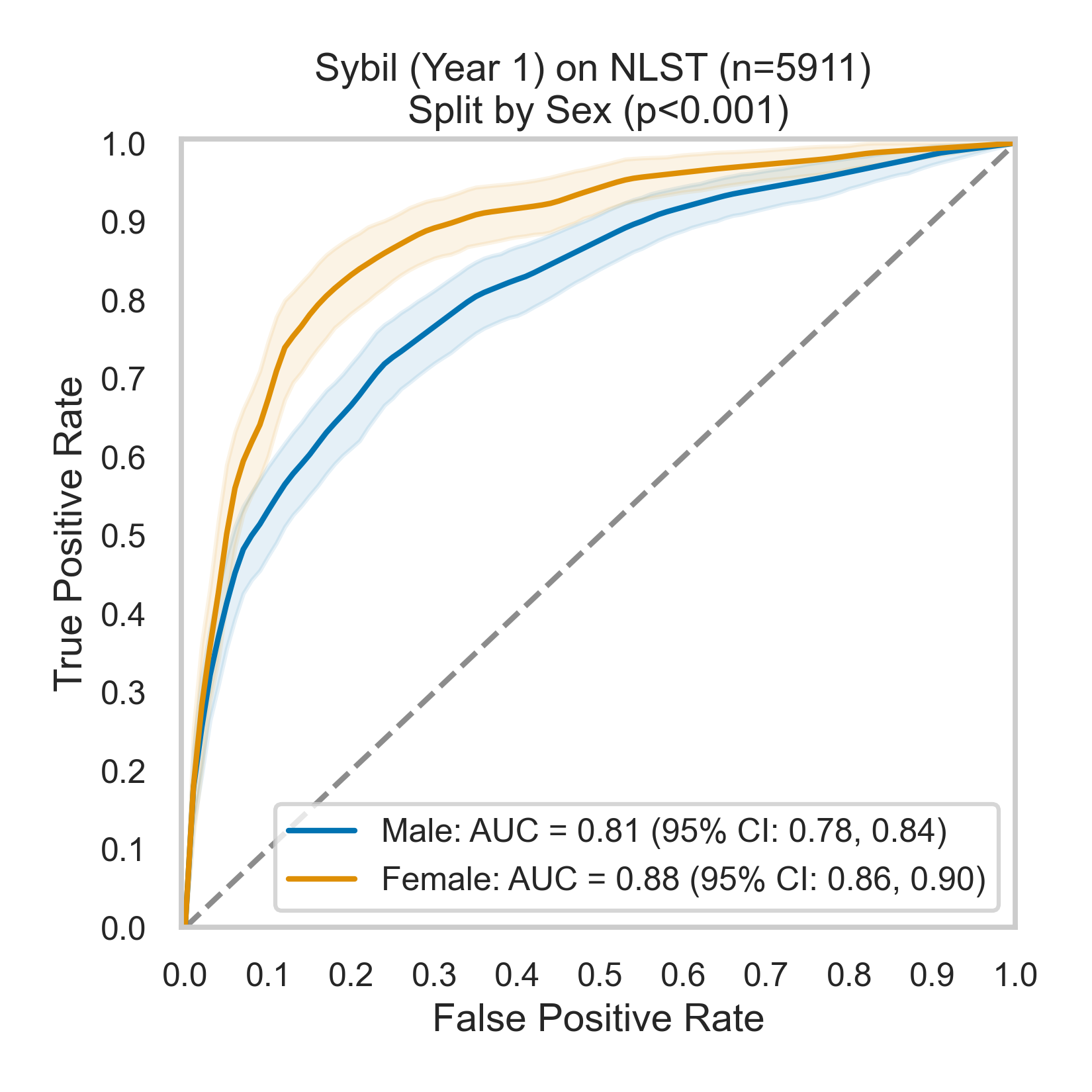}
    \caption{ROC curves (with 95\% CIs) for Sybil (Year 1) for men and women on NLST (n=5911 scans).}
    \label{fig:sybilROCgender5911}
\end{figure}

Sybil demonstrated a statistically significant difference in performance by sex, achieving an AUROC of 0.88 (0.86, 0.90) for women and 0.81 (0.78, 0.84) for men ($p < .001$), as shown in Figure \ref{fig:sybilROCgender5911}. Controlling for 90\% specificity overall, Sybil showed a substantial difference of 0.13 in sensitivity between women (Sensitivity: 0.66 (0.60, 0.71)) and men (Sensitivity: 0.53 (0.48, 0.58)), as shown in Table \ref{tab:resNLSTfullTPR}. Controlling for 90\% sensitivity, Sybil exhibited a 0.10 lower specificity for men (Specificity: 0.46 (0.44, 0.48)) than women (Specificity: 0.56 (0.54, 0.58)). The Venkadesh21 model showed a 0.04 higher specificity for women at the Brock ILST threshold, though no other substantial sensitivity and specificity disparities, and no significant AUROC disparity. Conversely, PanCan2b appeared to have slightly higher sensitivity for women than men, but substantially lower specificity across thresholds, up to 0.13 lower at a 90\% sensitivity threshold (see Table \ref{tab:resNLSTfullTNR}). 

In this study, Sybil performed non-significantly better for participants with a BMI of 25 $kg/m^2$ or above (AUROC: 0.86 (0.84, 0.88)) than those with a lower BMI (AUROC: 0.81 (0.78, 0.85), $p = 0.05$). PanCan2b performed significantly better for the higher-BMI subgroup (AUROC: 0.81 (0.79, 0.83)) than the lower-BMI subgroup (AUROC: 0.73 (0.70, 0.77), $p = 0.001$), as shown in Table \ref{tab:resNLSTscanROC}. Considering the factors behind BMI, Sybil demonstrated a significant performance disparity between participants with a height shorter than 68 inches (AUROC: 0.87 (0.85, 0.89)) than with taller participants (AUROC: 0.80 (0.77, 0.83), $p < 0.001$), but had little to no performance disparity ($p = 0.82$) between participants weighing more or fewer than 180 pounds. All models appeared to have lower specificity (and thus more false positives) with scans from participants with a lower BMI, as shown in Figure \ref{fig:allTNRbmi5911}.


\revision{
\begin{figure*}[htb!]
    \centering
    \begin{subfigure}[b]{0.38\textwidth}
        \centering
        \includegraphics[width=\textwidth]{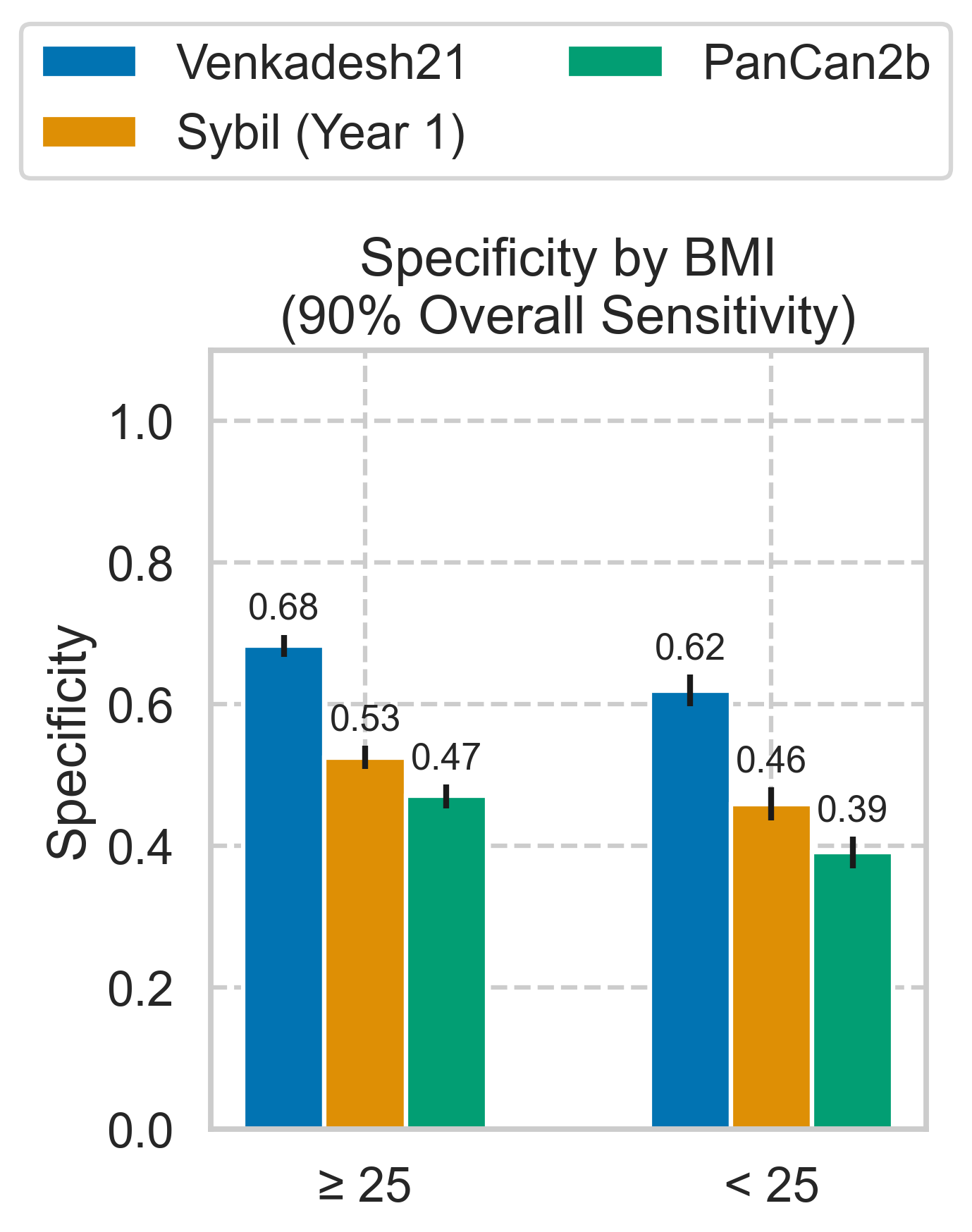}
        \caption{}
        \label{fig:allTNRbmi5911}
    \end{subfigure}
    \begin{subfigure}[b]{0.38\textwidth}
        \centering
        \includegraphics[width=\textwidth]{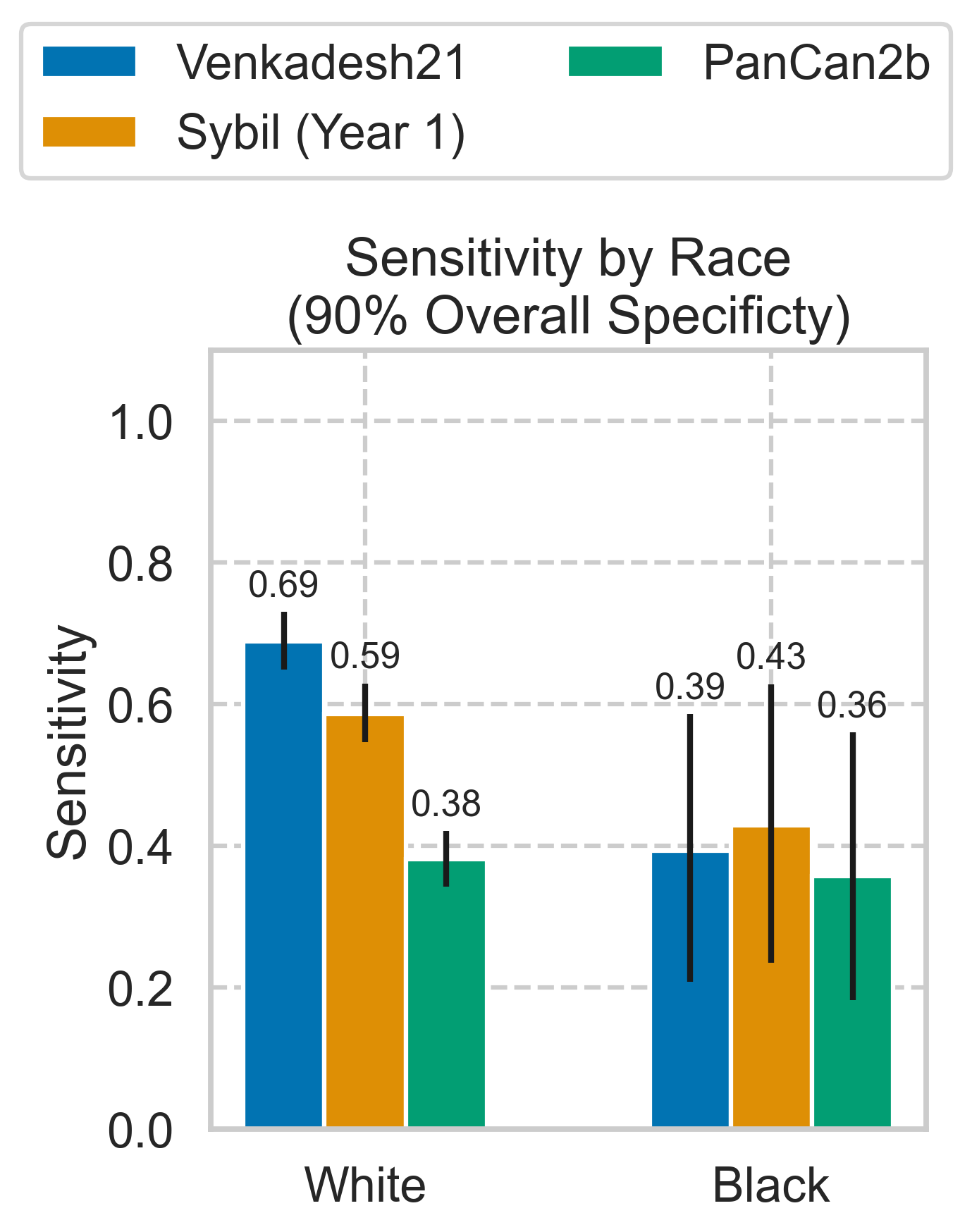}
        \caption{}
        \label{fig:allTPRrace5911}
    \end{subfigure}
    
    
    \begin{subfigure}[b]{0.38\textwidth}
        \centering
        \includegraphics[width=\textwidth]{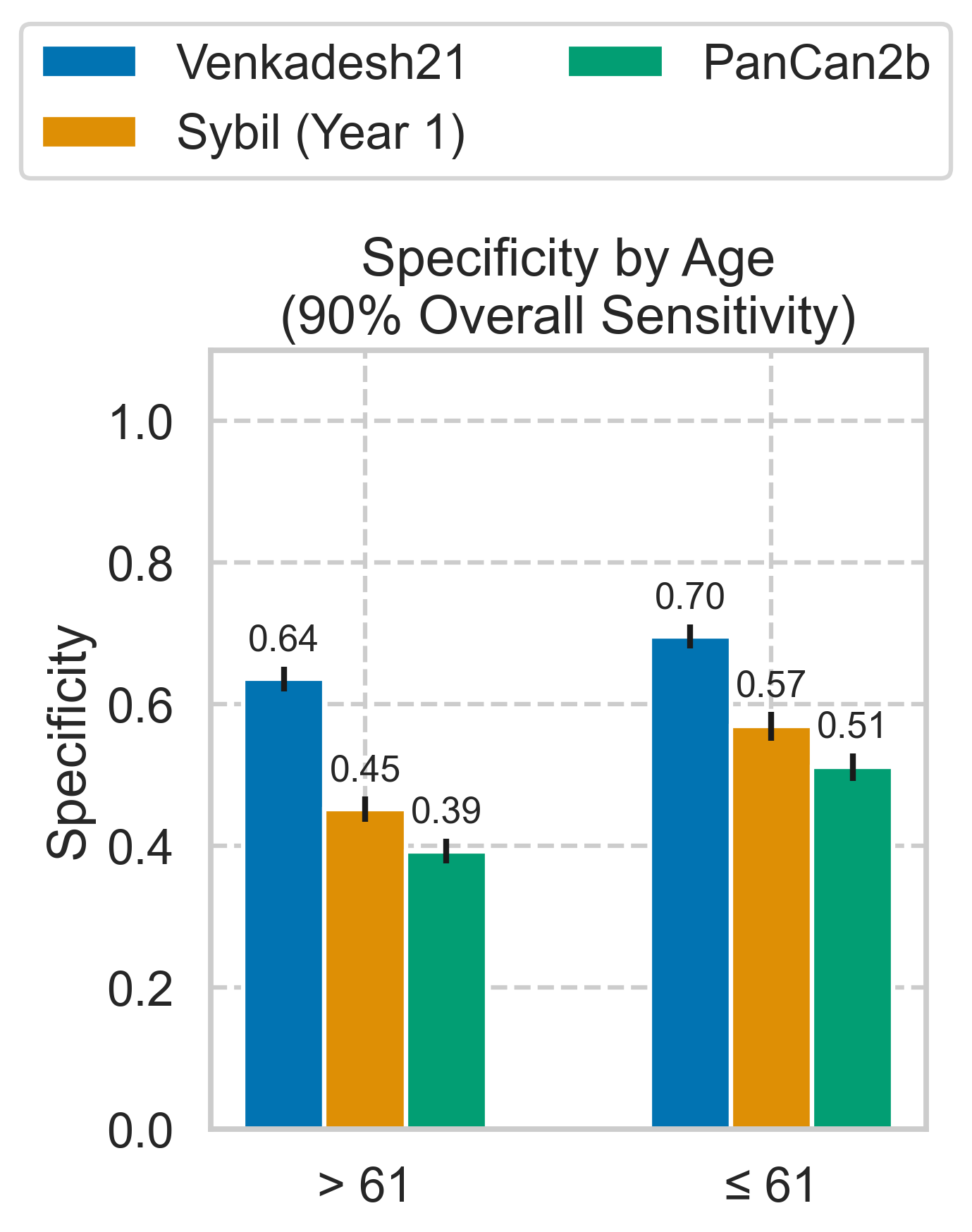}
        \caption{}
        \label{fig:allTNRage5911}
    \end{subfigure}
    \begin{subfigure}[b]{0.38\textwidth}
        \centering
        \includegraphics[width=\textwidth]{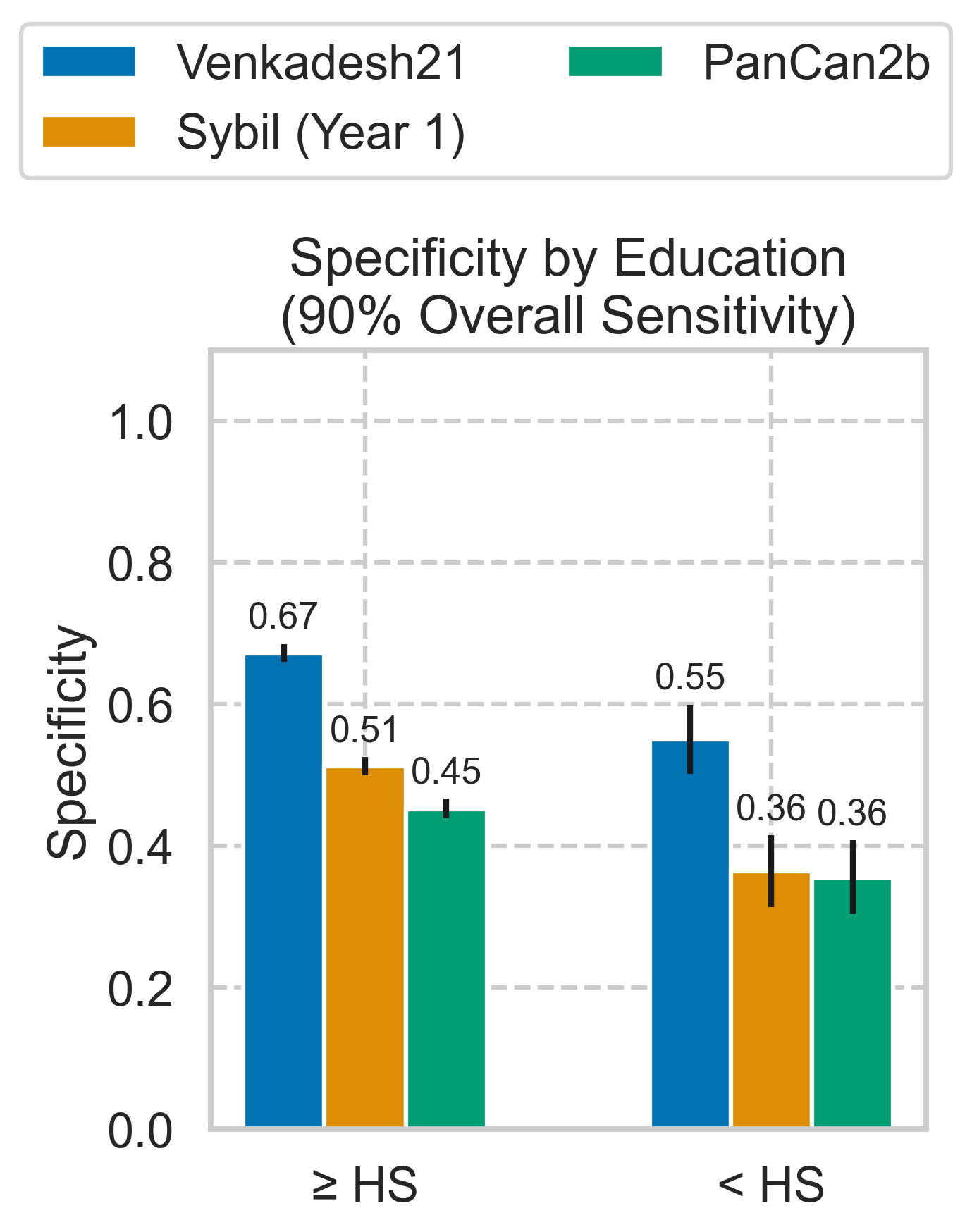}
        \caption{}
        \label{fig:allTNReducation5911}
    \end{subfigure}    
    \caption{Selected sensitivity and specificity of models on specific thresholds on NLST (n=5911 scans).}
    \label{fig:thresholdResults}
\end{figure*}
}


The Venkadesh21 model exhibited non-significantly better performance with White participants (AUROC: 0.89 (0.88, 0.91)) than Black participants (AUROC: 0.82 (0.74, 0.89), $p = 0.14$). When applying thresholds, we observed that this model had a higher sensitivity for White participants, as shown in Figure \ref{fig:allTPRrace5911}. In particular, the Venkadesh21 model had a substantial 0.30 difference in sensitivity for White participants (Sensitivity: 0.69 (0.65, 0.73)) than Black participants (Sensitivity: 0.39 (0.23, 0.59)) when applying a threshold which would result in a 10\% overall false positive rate (or 90\% specificity) for this model. Even without a significant AUROC disparity, this disparity indicates that the Venkadesh21 model has disparate performance between racial subgroups (see Figure \ref{fig:flowchart}).


Finally, we observed higher false positive rates from all three models with NLST participants older than 61 years of age, and those who had not graduated high school or received further education. While we did not observe significant differences in AUROC or specificity between participants with lower and higher education, all models achieved substantially lower specificity for participants who had not graduated high school in the 90\% sensitivity threshold (see Figure \ref{fig:allTNReducation5911}).
Between age groups, all models exhibited substantially lower specificity (0.07 lower for Venkadesh21 and 0.12 lower for the other models) for older participants when controlling sensitivity to 90\%, as shown in Figure \ref{fig:allTNRage5911}.


\begin{table*}[ht!]
\centering
\caption{Sensitivity (with 95\% confidence intervals) for models for demographic subgroups on NLST. Single asterisk (*) = sensitivity of one subgroup is outside the CI of the other. Double asterisks (**) = CIs do not intersect.}
\label{tab:resNLSTfullTPR}
\begin{tabular}{lll|ll|ll|ll}
\toprule
{} & {} & {} & \multicolumn{2}{c}{Venkadesh21} & \multicolumn{2}{c}{Sybil (Year 1)} & \multicolumn{2}{c}{PanCan2b} \\
{Policy} & {Attribute} & {Group} & {Sensitivity} & {CI} & {Sensitivity} & {CI} & {Sensitivity} & {CI} \\
\midrule
\multirow[c]{14}{*}{\rotatebox[origin=c]{90}{90\% \revision{Overall} Sensitivity}} & \multirow[c]{2}{*}{Age} & $>$ 61 & 0.90 (0.86, 0.93) &  & 0.91 (0.88, 0.94) &  & 0.92 (0.89, 0.95) &  \\
 &  & $\leqslant$ 61 & 0.91 (0.87, 0.94) &  & 0.90 (0.86, 0.94) &  & 0.87 (0.82, 0.91) & * \\
\cline{2-9}
 & \multirow[c]{2}{*}{BMI} & $\geqslant$ 25 & 0.90 (0.87, 0.93) &  & 0.92 (0.89, 0.94) &  & 0.91 (0.88, 0.94) &  \\
 &  & $<$ 25 & 0.90 (0.86, 0.94) &  & 0.89 (0.84, 0.93) &  & 0.88 (0.84, 0.92) &  \\
\cline{2-9}
 & \multirow[c]{2}{*}{Education} & $\geqslant$ HS & 0.90 (0.87, 0.92) &  & 0.91 (0.88, 0.93) &  & 0.90 (0.88, 0.93) &  \\
 &  & $<$ HS & 0.88 (0.77, 0.97) &  & 0.90 (0.80, 0.98) &  & 0.88 (0.78, 0.97) &  \\
\cline{2-9}
 & \multirow[c]{2}{*}{Height} & $\leqslant$ 68 & 0.90 (0.87, 0.94) &  & 0.92 (0.90, 0.95) &  & 0.92 (0.89, 0.95) &  \\
 &  & $>$ 68 & 0.90 (0.86, 0.93) &  & 0.89 (0.85, 0.92) &  & 0.88 (0.84, 0.92) & * \\
\cline{2-9}
 & \multirow[c]{2}{*}{Race} & White & 0.90 (0.88, 0.93) &  & 0.91 (0.89, 0.93) &  & 0.90 (0.87, 0.93) &  \\
 &  & Black & 0.82 (0.67, 0.96) &  & 0.86 (0.71, 0.97) &  & 0.89 (0.75, 1.00) &  \\
\cline{2-9}
 & \multirow[c]{2}{*}{Sex} & Male & 0.90 (0.87, 0.93) &  & 0.90 (0.86, 0.93) &  & 0.88 (0.85, 0.92) &  \\
 &  & Female & 0.90 (0.86, 0.93) &  & 0.92 (0.89, 0.96) &  & 0.92 (0.89, 0.95) & * \\
\cline{2-9}
 & \multirow[c]{2}{*}{Weight} & $\leqslant$ 180 & 0.91 (0.88, 0.94) &  & 0.92 (0.89, 0.95) &  & 0.91 (0.88, 0.94) &  \\
 &  & $>$ 180 & 0.89 (0.85, 0.93) &  & 0.89 (0.86, 0.93) &  & 0.89 (0.85, 0.92) &  \\
\cline{1-9} \cline{2-9}
\multirow[c]{14}{*}{\rotatebox[origin=c]{90}{90\% \revision{Overall} Specificity}} & \multirow[c]{2}{*}{Age} & $>$ 61 & 0.70 (0.65, 0.75) &  & 0.61 (0.56, 0.66) &  & 0.40 (0.35, 0.44) &  \\
 &  & $\leqslant$ 61 & 0.65 (0.58, 0.71) &  & 0.53 (0.47, 0.60) & * & 0.37 (0.32, 0.44) &  \\
\cline{2-9}
 & \multirow[c]{2}{*}{BMI} & $\geqslant$ 25 & 0.68 (0.63, 0.72) &  & 0.58 (0.53, 0.62) &  & 0.37 (0.33, 0.43) &  \\
 &  & $<$ 25 & 0.69 (0.63, 0.75) &  & 0.59 (0.52, 0.65) &  & 0.41 (0.34, 0.48) &  \\
\cline{2-9}
 & \multirow[c]{2}{*}{Education} & $\geqslant$ HS & 0.68 (0.64, 0.72) &  & 0.59 (0.55, 0.63) &  & 0.39 (0.35, 0.43) &  \\
 &  & $<$ HS & 0.68 (0.54, 0.82) &  & 0.56 (0.41, 0.71) &  & 0.41 (0.26, 0.57) &  \\
\cline{2-9}
 & \multirow[c]{2}{*}{Height} & $\leqslant$ 68 & 0.70 (0.65, 0.75) &  & 0.63 (0.58, 0.68) &  & 0.43 (0.38, 0.48) &  \\
 &  & $>$ 68 & 0.66 (0.61, 0.71) &  & 0.52 (0.47, 0.58) & * & 0.34 (0.29, 0.40) & * \\
\cline{2-9}
 & \multirow[c]{2}{*}{Race} & White & 0.69 (0.65, 0.73) &  & 0.59 (0.55, 0.63) &  & 0.38 (0.34, 0.42) &  \\
 &  & Black & 0.39 (0.21, 0.59) & ** & 0.43 (0.24, 0.63) &  & 0.36 (0.18, 0.56) &  \\
\cline{2-9}
 & \multirow[c]{2}{*}{Sex} & Male & 0.66 (0.61, 0.71) &  & 0.53 (0.48, 0.58) &  & 0.33 (0.28, 0.38) &  \\
 &  & Female & 0.71 (0.65, 0.76) &  & 0.66 (0.60, 0.71) & ** & 0.47 (0.41, 0.53) & ** \\
\cline{2-9}
 & \multirow[c]{2}{*}{Weight} & $\leqslant$ 180 & 0.69 (0.64, 0.74) &  & 0.59 (0.54, 0.64) &  & 0.42 (0.36, 0.47) &  \\
 &  & $>$ 180 & 0.68 (0.62, 0.74) &  & 0.57 (0.51, 0.63) &  & 0.35 (0.29, 0.41) & * \\
\cline{1-9} \cline{2-9}
\multirow[c]{14}{*}{\rotatebox[origin=c]{90}{Brock ILST (6\%)}} & \multirow[c]{2}{*}{Age} & $>$ 61 & 0.88 (0.84, 0.91) &  & 0.61 (0.56, 0.66) &  & 0.72 (0.66, 0.76) &  \\
 &  & $\leqslant$ 61 & 0.89 (0.85, 0.93) &  & 0.53 (0.47, 0.60) & * & 0.67 (0.61, 0.73) &  \\
\cline{2-9}
 & \multirow[c]{2}{*}{BMI} & $\geqslant$ 25 & 0.89 (0.86, 0.92) &  & 0.58 (0.53, 0.62) &  & 0.70 (0.65, 0.75) &  \\
 &  & $<$ 25 & 0.88 (0.82, 0.92) &  & 0.59 (0.52, 0.65) &  & 0.69 (0.63, 0.76) &  \\
\cline{2-9}
 & \multirow[c]{2}{*}{Education} & $\geqslant$ HS & 0.88 (0.85, 0.91) &  & 0.59 (0.55, 0.63) &  & 0.69 (0.65, 0.73) &  \\
 &  & $<$ HS & 0.88 (0.77, 0.97) &  & 0.56 (0.41, 0.71) &  & 0.80 (0.68, 0.92) &  \\
\cline{2-9}
 & \multirow[c]{2}{*}{Height} & $\leqslant$ 68 & 0.89 (0.85, 0.92) &  & 0.63 (0.58, 0.68) &  & 0.72 (0.67, 0.77) &  \\
 &  & $>$ 68 & 0.88 (0.84, 0.91) &  & 0.52 (0.47, 0.58) & * & 0.68 (0.62, 0.73) &  \\
\cline{2-9}
 & \multirow[c]{2}{*}{Race} & White & 0.89 (0.86, 0.91) &  & 0.59 (0.55, 0.63) &  & 0.69 (0.65, 0.73) &  \\
 &  & Black & 0.75 (0.59, 0.91) &  & 0.43 (0.24, 0.63) &  & 0.68 (0.50, 0.85) &  \\
\cline{2-9}
 & \multirow[c]{2}{*}{Sex} & Male & 0.88 (0.84, 0.92) &  & 0.53 (0.48, 0.58) &  & 0.68 (0.62, 0.72) &  \\
 &  & Female & 0.89 (0.85, 0.92) &  & 0.66 (0.60, 0.71) & ** & 0.73 (0.68, 0.78) &  \\
\cline{2-9}
 & \multirow[c]{2}{*}{Weight} & $\leqslant$ 180 & 0.88 (0.84, 0.91) &  & 0.59 (0.54, 0.64) &  & 0.70 (0.65, 0.75) &  \\
 &  & $>$ 180 & 0.89 (0.85, 0.92) &  & 0.57 (0.51, 0.63) &  & 0.69 (0.64, 0.75) &  \\
\cline{1-9} \cline{2-9}
\bottomrule
\end{tabular}
\end{table*}

\begin{table*}[ht!]
\centering
\caption{Specificity (with 95\% confidence intervals) for models for demographic subgroups on NLST. Single asterisk (*) = specificity of one subgroup is outside the CI of the other. Double asterisks (**) = CIs do not intersect.}
\label{tab:resNLSTfullTNR}
\begin{tabular}{lll|ll|ll|ll}
\toprule
{} & {} & {} & \multicolumn{2}{c}{Venkadesh21} & \multicolumn{2}{c}{Sybil (Year 1)} & \multicolumn{2}{c}{PanCan2b} \\
{Policy} & {Attribute} & {Group} & {Specificity} & {CI} & {Specificity} & {CI} & {Specificity} & {CI} \\
\midrule
\multirow[c]{14}{*}{\rotatebox[origin=c]{90}{90\% \revision{Overall} Sensitivity}} & \multirow[c]{2}{*}{Age} & $>$ 61 & 0.64 (0.62, 0.65) &  & 0.45 (0.43, 0.47) &  & 0.39 (0.38, 0.41) &  \\
 &  & $\leqslant$ 61 & 0.70 (0.68, 0.71) & ** & 0.57 (0.55, 0.59) & ** & 0.51 (0.49, 0.53) & ** \\
\cline{2-9}
 & \multirow[c]{2}{*}{BMI} & $\geqslant$ 25 & 0.68 (0.67, 0.70) &  & 0.53 (0.51, 0.54) &  & 0.47 (0.45, 0.49) &  \\
 &  & $<$ 25 & 0.62 (0.60, 0.64) & ** & 0.46 (0.44, 0.48) & ** & 0.39 (0.37, 0.41) & ** \\
\cline{2-9}
 & \multirow[c]{2}{*}{Education} & $\geqslant$ HS & 0.67 (0.66, 0.69) &  & 0.51 (0.50, 0.53) &  & 0.45 (0.44, 0.47) &  \\
 &  & $<$ HS & 0.55 (0.50, 0.60) & ** & 0.36 (0.31, 0.41) & ** & 0.36 (0.30, 0.41) & ** \\
\cline{2-9}
 & \multirow[c]{2}{*}{Height} & $\leqslant$ 68 & 0.68 (0.66, 0.70) &  & 0.55 (0.53, 0.57) &  & 0.41 (0.39, 0.43) &  \\
 &  & $>$ 68 & 0.64 (0.63, 0.66) & * & 0.45 (0.43, 0.47) & ** & 0.48 (0.46, 0.50) & ** \\
\cline{2-9}
 & \multirow[c]{2}{*}{Race} & White & 0.66 (0.65, 0.67) &  & 0.50 (0.49, 0.51) &  & 0.45 (0.43, 0.46) &  \\
 &  & Black & 0.73 (0.66, 0.80) &  & 0.59 (0.52, 0.66) & ** & 0.37 (0.29, 0.44) & * \\
\cline{2-9}
 & \multirow[c]{2}{*}{Sex} & Male & 0.65 (0.63, 0.66) &  & 0.46 (0.45, 0.48) &  & 0.50 (0.48, 0.52) &  \\
 &  & Female & 0.68 (0.66, 0.70) & * & 0.56 (0.54, 0.58) & ** & 0.37 (0.35, 0.39) & ** \\
\cline{2-9}
 & \multirow[c]{2}{*}{Weight} & $\leqslant$ 180 & 0.65 (0.63, 0.67) &  & 0.49 (0.47, 0.51) &  & 0.40 (0.38, 0.42) &  \\
 &  & $>$ 180 & 0.68 (0.66, 0.69) & * & 0.51 (0.49, 0.53) &  & 0.49 (0.47, 0.51) & ** \\
\cline{1-9} \cline{2-9}
\multirow[c]{14}{*}{\rotatebox[origin=c]{90}{90\% \revision{Overall} Specificity}} & \multirow[c]{2}{*}{Age} & $>$ 61 & 0.88 (0.87, 0.89) &  & 0.89 (0.88, 0.90) &  & 0.88 (0.86, 0.89) &  \\
 &  & $\leqslant$ 61 & 0.92 (0.91, 0.93) & ** & 0.92 (0.91, 0.93) & ** & 0.93 (0.92, 0.94) & ** \\
\cline{2-9}
 & \multirow[c]{2}{*}{BMI} & $\geqslant$ 25 & 0.91 (0.90, 0.92) &  & 0.91 (0.90, 0.92) &  & 0.92 (0.91, 0.92) &  \\
 &  & $<$ 25 & 0.87 (0.86, 0.89) & ** & 0.89 (0.87, 0.90) & ** & 0.87 (0.85, 0.88) & ** \\
\cline{2-9}
 & \multirow[c]{2}{*}{Education} & $\geqslant$ HS & 0.90 (0.89, 0.91) &  & 0.90 (0.90, 0.91) &  & 0.90 (0.89, 0.91) &  \\
 &  & $<$ HS & 0.86 (0.83, 0.90) & * & 0.89 (0.86, 0.92) &  & 0.88 (0.84, 0.91) &  \\
\cline{2-9}
 & \multirow[c]{2}{*}{Height} & $\leqslant$ 68 & 0.90 (0.89, 0.91) &  & 0.90 (0.89, 0.92) &  & 0.90 (0.88, 0.91) &  \\
 &  & $>$ 68 & 0.90 (0.89, 0.92) &  & 0.90 (0.89, 0.91) &  & 0.91 (0.90, 0.92) &  \\
\cline{2-9}
 & \multirow[c]{2}{*}{Race} & White & 0.90 (0.89, 0.91) &  & 0.90 (0.90, 0.91) &  & 0.90 (0.89, 0.91) &  \\
 &  & Black & 0.93 (0.89, 0.97) &  & 0.92 (0.88, 0.96) &  & 0.88 (0.83, 0.93) &  \\
\cline{2-9}
 & \multirow[c]{2}{*}{Sex} & Male & 0.90 (0.89, 0.91) &  & 0.90 (0.89, 0.91) &  & 0.91 (0.90, 0.92) &  \\
 &  & Female & 0.90 (0.88, 0.91) &  & 0.90 (0.89, 0.91) &  & 0.88 (0.87, 0.89) & ** \\
\cline{2-9}
 & \multirow[c]{2}{*}{Weight} & $\leqslant$ 180 & 0.89 (0.88, 0.90) &  & 0.89 (0.88, 0.91) &  & 0.88 (0.87, 0.89) &  \\
 &  & $>$ 180 & 0.91 (0.90, 0.92) & ** & 0.91 (0.90, 0.92) & * & 0.92 (0.91, 0.93) & ** \\
\cline{1-9} \cline{2-9}
\multirow[c]{14}{*}{\rotatebox[origin=c]{90}{Brock ILST (6\%)}} & \multirow[c]{2}{*}{Age} & $>$ 61 & 0.67 (0.65, 0.69) &  & 0.89 (0.88, 0.90) &  & 0.69 (0.68, 0.71) &  \\
 &  & $\leqslant$ 61 & 0.73 (0.71, 0.75) & ** & 0.92 (0.91, 0.93) & ** & 0.79 (0.77, 0.81) & ** \\
\cline{2-9}
 & \multirow[c]{2}{*}{BMI} & $\geqslant$ 25 & 0.72 (0.70, 0.73) &  & 0.91 (0.90, 0.92) &  & 0.76 (0.75, 0.77) &  \\
 &  & $<$ 25 & 0.66 (0.63, 0.68) & ** & 0.89 (0.87, 0.90) & ** & 0.68 (0.66, 0.71) & ** \\
\cline{2-9}
 & \multirow[c]{2}{*}{Education} & $\geqslant$ HS & 0.71 (0.69, 0.72) &  & 0.90 (0.90, 0.91) &  & 0.74 (0.73, 0.75) &  \\
 &  & $<$ HS & 0.59 (0.55, 0.64) & ** & 0.89 (0.86, 0.92) &  & 0.69 (0.64, 0.73) & * \\
\cline{2-9}
 & \multirow[c]{2}{*}{Height} & $\leqslant$ 68 & 0.71 (0.70, 0.73) &  & 0.90 (0.89, 0.92) &  & 0.72 (0.70, 0.73) &  \\
 &  & $>$ 68 & 0.68 (0.66, 0.69) & ** & 0.90 (0.89, 0.91) &  & 0.76 (0.74, 0.77) & ** \\
\cline{2-9}
 & \multirow[c]{2}{*}{Race} & White & 0.69 (0.68, 0.71) &  & 0.90 (0.90, 0.91) &  & 0.74 (0.72, 0.75) &  \\
 &  & Black & 0.76 (0.68, 0.82) &  & 0.92 (0.88, 0.96) &  & 0.68 (0.60, 0.75) &  \\
\cline{2-9}
 & \multirow[c]{2}{*}{Sex} & Male & 0.68 (0.66, 0.70) &  & 0.90 (0.89, 0.91) &  & 0.77 (0.76, 0.79) &  \\
 &  & Female & 0.72 (0.70, 0.74) & ** & 0.90 (0.89, 0.91) &  & 0.69 (0.67, 0.71) & ** \\
\cline{2-9}
 & \multirow[c]{2}{*}{Weight} & $\leqslant$ 180 & 0.69 (0.67, 0.70) &  & 0.89 (0.88, 0.91) &  & 0.69 (0.68, 0.71) &  \\
 &  & $>$ 180 & 0.71 (0.69, 0.72) & * & 0.91 (0.90, 0.92) & * & 0.78 (0.76, 0.80) & ** \\
\cline{1-9} \cline{2-9}
\bottomrule
\end{tabular}
\end{table*}

\subsection{Fairness Assessment}

In this analysis, we examined whether the observed performance biases could be explained by differences in the prevalence of clinical risk factors. We investigated Sybil’s performance disparity between men and women, the Venkadesh21 model’s lower sensitivity for Black participants compared to White participants at a 90\% specificity threshold, AUROC and specificity disparities observed in Sybil and PanCan2b between participants with a low and high BMI, and the disparate specificity for all models between participants who had and had not graduated high school. We did not separately assess confounders for Sybil's disparity between height-based subgroups, since it is directly a factor in BMI, and highly related to sex in the NLST cohort. We also did not assess potential confounding for the apparent overestimation of lung cancer risk in older LCS participants, as age is a clinically established risk factor for developing cancer \citep{deGroot2018-EducationSmoking}. Tables displaying comprehensive results are provided in the Appendix.

\subsubsection{Disparity for Sybil Between Sexes}

We first assess Sybil's disparity between sexes. Men in the NLST cohort were more likely to smoke pipes and cigars and work in a hazardous field for the lungs without a mask, while women were more likely to live with a smoker and have a previous pneumonia diagnosis. Table \ref{tab:genderIsolationPlusROC} (see Appendix) displays the 10 characteristics with the greatest prevalence differences between men and women. Men had higher smoking rates in NLST, with a median of 55 pack-years (IQR: 35) compared to 46 pack-years (IQR: 22) for women, though with similar rates of lung cancer (10\% for both, as seen in Table \ref{tab:datasetDemos}). Male participants' median weight was 35 pounds heavier (and median height 7 inches taller) than that of women, but median BMIs were similar.

\begin{figure*}[htb!]
    \centering
    \includegraphics[width=0.9\linewidth]{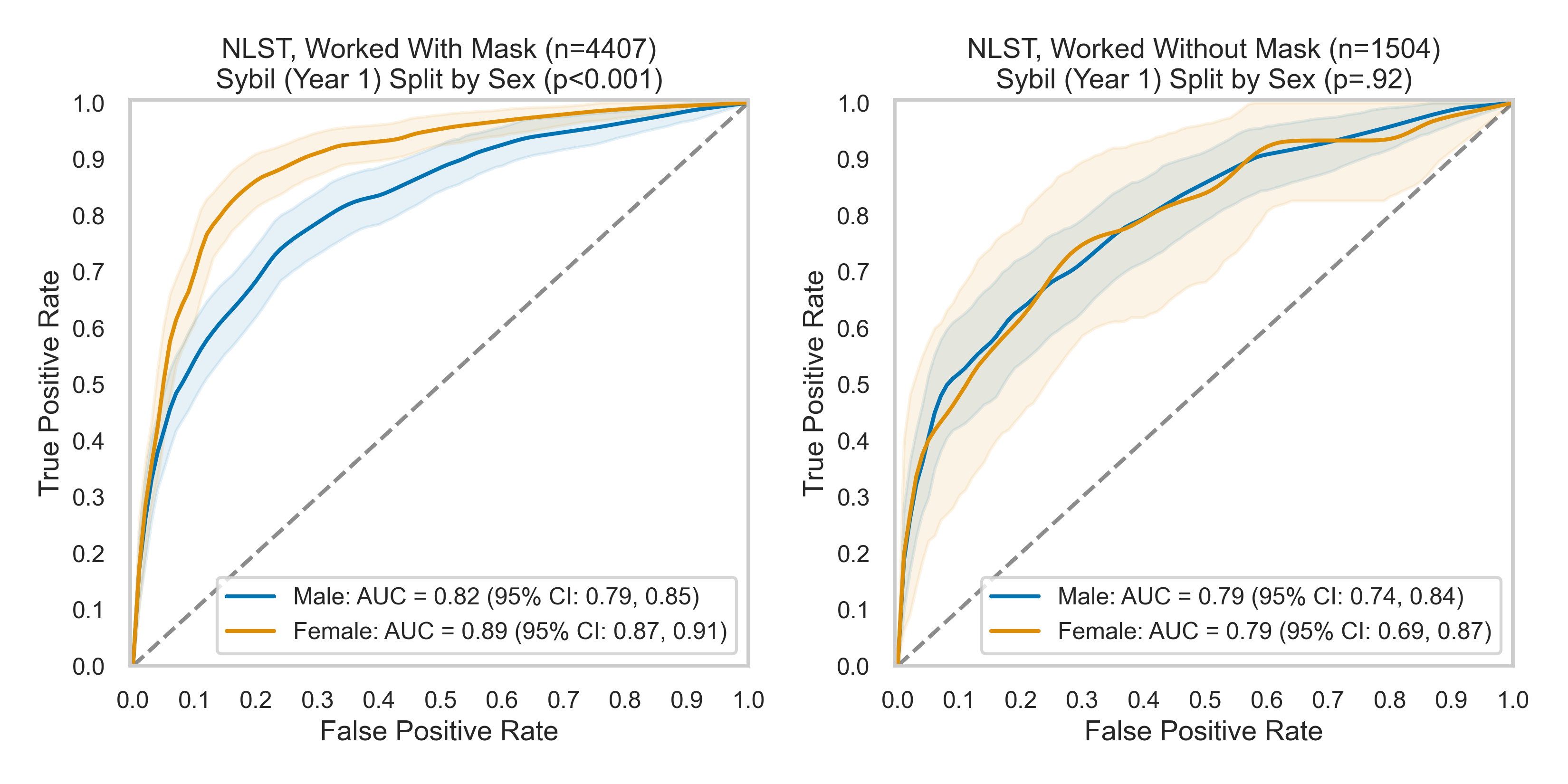}
    \caption{ROC curves (with 95\% CIs) for Sybil (Year 1) between men and women when isolating for a self-reported work history of working in a dangerous occupation for the lungs with or without a mask.}
    \label{fig:sybilROCgender-WrkNoMask}
\end{figure*}

For almost all of these factors, Sybil's AUROC disparity between men and women persisted when isolating for (or excluding) any single condition, as shown in Table \ref{tab:genderIsolationPlusROC} in the Appendix. As an example, men had a 34\% rate of working in a dangerous field without a mask, while women only had a 13\% rate of doing so. Isolating for only participants who worked without a mask, men (109 malignant scans, 1075 benign scans) and women (30 malignant scans, 290 benign scans) both had an AUROC of 0.79 ($p = 0.92$), as shown in Figure \ref{fig:sybilROCgender-WrkNoMask}. However, for participants without this work history, Sybil still performed significantly better for women (AUROC: 0.89 (0.87, 0.91)) than for men (AUROC: 0.82 (0.79, 0.85), $p < 0.001$). This indicates that working without a mask did not confound Sybil's sex bias, since the disparity remained when excluding for working without a mask (see Figure \ref{fig:flowchart}). Conversely, the AUROC disparity between men and women was no longer significant when splitting the NLST cohort based on height. While there was still a 0.04 increase in AUROC from men to women, the difference was not significant for participants taller than 68 inches ($p = 0.63$) or for shorter participants ($p = 0.17$). This was also the factor with the greatest prevalence disparity between men and women, as 76\% of men in NLST have a height greater than 68 inches, while this is the case for only 6\% of women.

We also examined the impact of these factors on explaining the substantial disparity between sexes in sensitivity (at 90\% specificity) and specificity (at 90\% sensitivity). Examining these metrics revealed that Sybil still maintained performance disparities between men and women, even when isolating for potential confounders (see Table \ref{tab:genderTPRandTNRisolated} in the Appendix). For each of these characteristics, Sybil's sensitivity and specificity was substantially better for women, with almost no CIs intersecting between sexes for any comparisons. Sybil demonstrated 0.08 better sensitivity for taller women than taller men and 0.11 greater sensitivity for shorter women than shorter men, with CIs intersecting between sexes. However, while the specificity for taller men and women were the same, shorter men tended to receive a substantially higher false positive rate than shorter women, with a gap of 0.08 and non-intersecting CIs.

\subsubsection{Racial Disparity for Venkadesh21}

With race, we focused on investigating the Venkadesh21 model's substantial sensitivity disparity between White and Black participants at 90\% specificity, along with its non-significant AUROC disparity. White participants accounted for 93\% of the NLST cohort, while Black participants accounted for only 3\% of the data. Despite White participants having higher smoking rates (with medians of 51 pack-years and 25 average cigarettes per day) in NLST than Black participants (with medians of 43 pack-years and 20 average cigarettes per day), Black participants had a higher rate of lung cancer in the NLST cohort (15\%) than White participants (10\%). Black participants had higher rates of being current smokers (70\%) instead of former smokers than White participants (50\%), and had a previous hypertension diagnosis at a much higher rate (54\%) than White participants (33\%), as shown in Table \ref{tab:raceIsolationPlusROC} in the Appendix.

\begin{figure*}[htb!]
    \centering
    \includegraphics[width=0.9\linewidth]{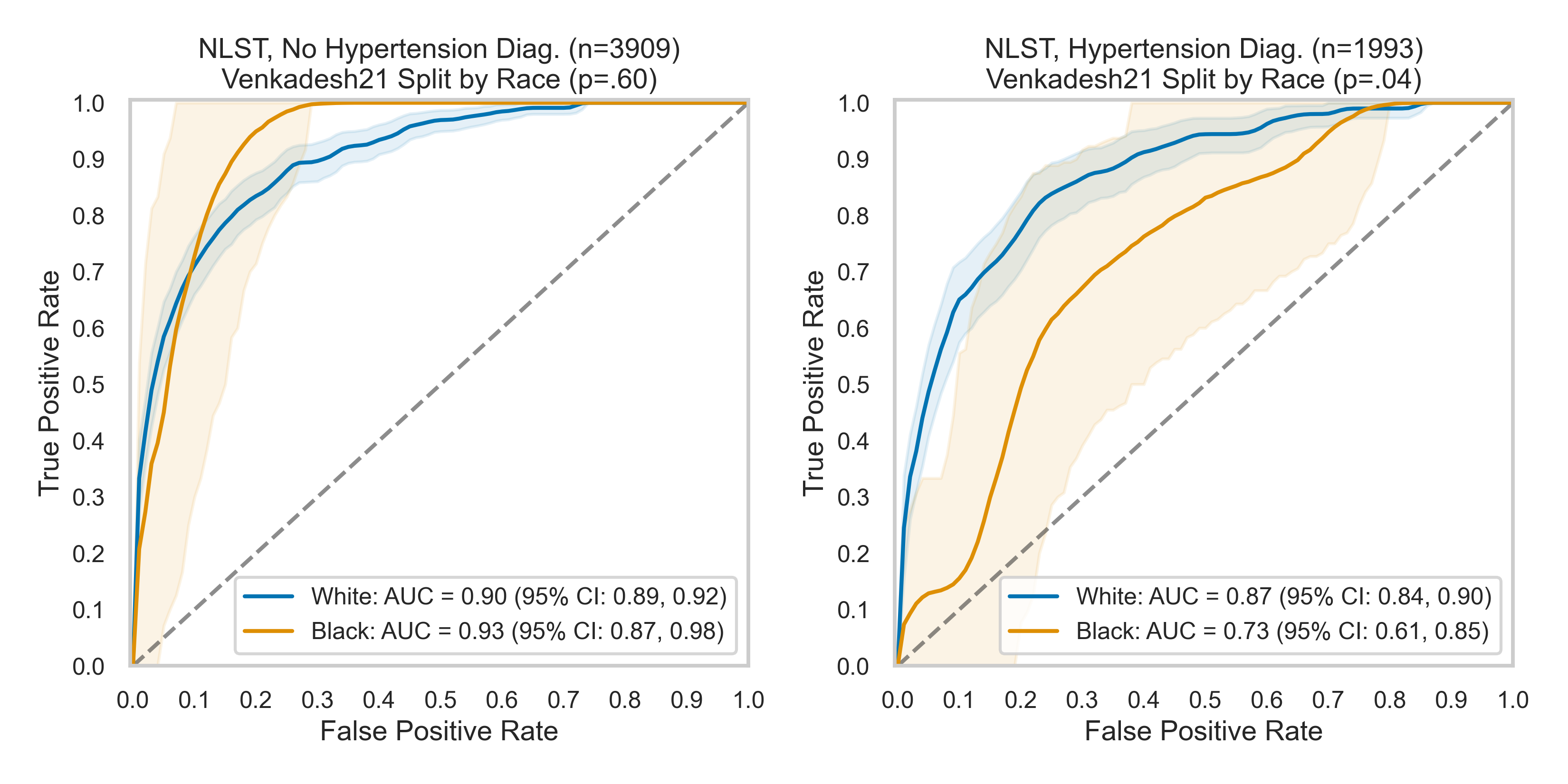}
    \caption{ROC curves (with 95\% CIs) for Venkadesh21 between racial groups when isolating for a hypertension diagnosis.}
    \label{fig:venkROCrace-Hypertension}
\end{figure*}

Investigating the impact of these confounders revealed that the Venkadesh21 model still exhibited (non-significantly) worse AUROC scores (see Table \ref{tab:raceIsolationPlusROC} in the Appendix) and substantially lower sensitivity (see Table \ref{tab:raceTPRandTNRisolated} in the Appendix) for Black participants. As an example, we investigated if the greater hypertension prevalence in Black participants confounded the Venkadesh21 model's reduced performance. When plotting ROC curves for participants without hypertension, we observed similar performance for Black participants (AUROC: 0.93, (0.87, 0.98)) than White participants (AUROC: 0.90 (0.89, 0.92), $p = 0.60$). However, we did observe an AUROC performance difference between White participants with hypertension (196 malignant scans, 1617 benign scans, AUROC: 0.87 (0.84, 0.90)) and Black participants with hypertension (16 malignant scans, 86 benign scans, AUROC: 0.73 (0.61, 0.85), $p = 0.04$), as shown in Figure \ref{fig:venkROCrace-Hypertension}. The sensitivity disparity, shown in Table \ref{tab:raceTPRandTNRisolated} (see Appendix), also remained substantial between White participants with a hypertension diagnosis (Sensitivity: 0.65 (0.59, 0.72)) and Black participants with a hypertension diagnosis (Sensitivity: 0.19 (0.00, 0.40)). This indicates that a prevalence disparity for hypertension did not confound lower sensitivity for Black participants, as the Venkadesh21 model still showed disparate performance when isolating for participants diagnosed with this condition. 

\subsubsection{Body Mass Index Disparity}

We also investigated further the lower AUROC and higher false positive rates for participants with a lower body mass index from our models, in particular Sybil's Year 1 risk score. Participants with a lower BMI in NLST were more likely to be current (and not former) smokers, were more likely to be pipe smokers, and had smoked more cigarettes per day than high BMI participants, but had similar pack-years, as detailed in Table \ref{tab:bmiIsolationPlusROC} in the Appendix. Lower-BMI participants were more likely to have emphysema in \revision{a} LDCT (47\% compared to 32\% for high-BMI participants) or be previously diagnosed with it, while high-BMI participants were more likely to have a diabetes or hypertension diagnosis. 

\begin{figure*}[htb!]
    \centering
    \includegraphics[width=0.9\linewidth]{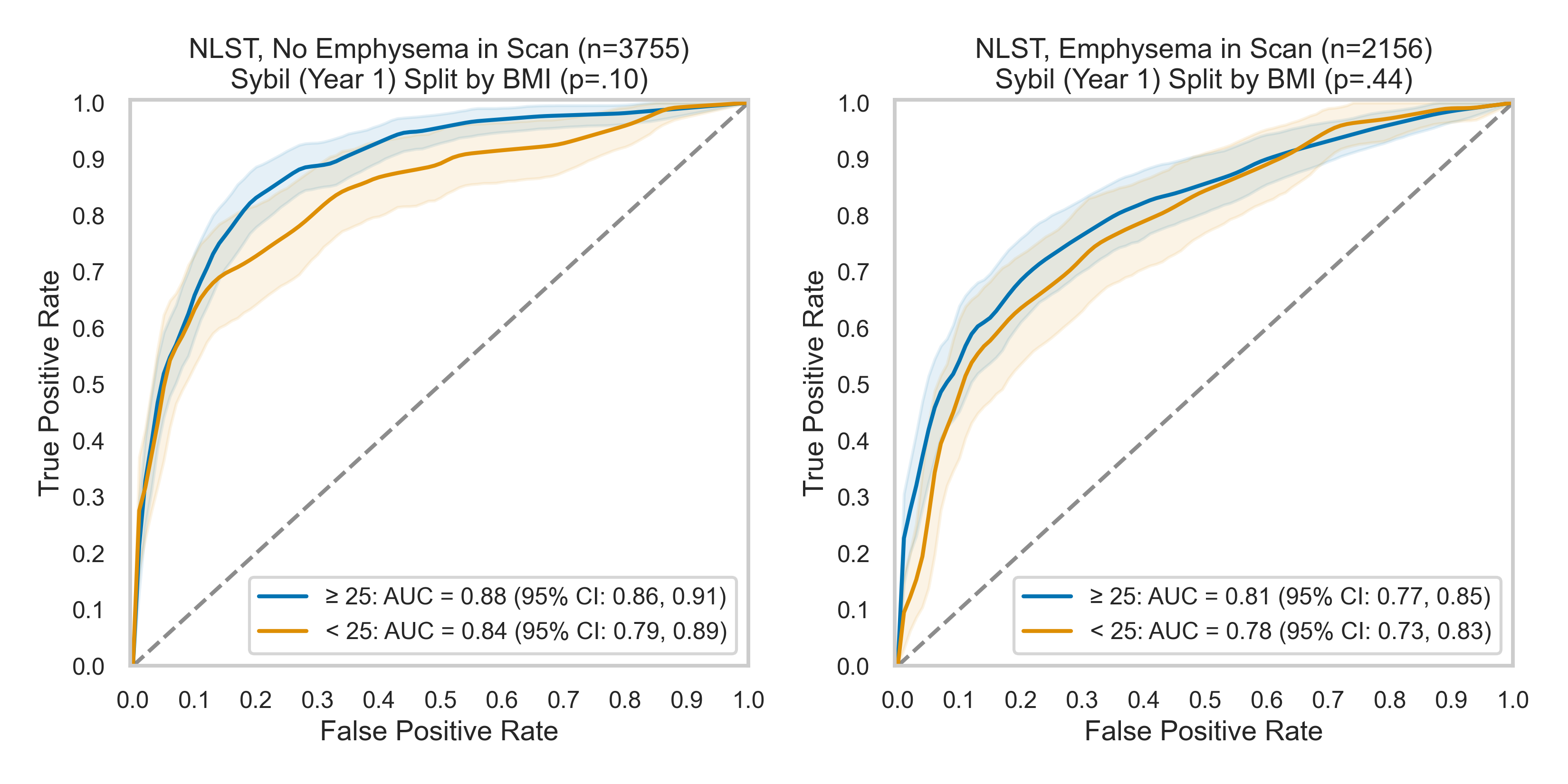}
    \caption{ROC curves (with 95\% CIs) for Sybil (Year 1) between BMI groups when isolating for emphysema.}
    \label{fig:sybilROCbmi-Emphysema}
\end{figure*}

Disparities persisted for most characteristics between high and low BMI participants. However, emphysema in the LDCT may have confounded Sybil's AUROC and specificity disparities, since isolating and excluding this factor resulted in reduced disparities between subgroups. Figure \ref{fig:sybilROCbmi-Emphysema} shows that for low-BMI participants with emphysema in their scans (103 malignant, 789 benign), Sybil's AUROC was 0.78 (0.73, 0.83), close to that for high-BMI participants (161 malignant scans, 1103 benign scans, AUROC: 0.81 (0.77, 0.85), $p = 0.44$). Sybil exhibited similar performance for participants without emphysema. The specificity disparity was also smaller when isolating for emphysema (see Table \ref{tab:bmiTPRandTNRisolated} in the Appendix). Sybil's lower performance for low-BMI participants aligned with its significantly worse ($p < 0.001$) performance with emphysema (see Figure \ref{fig:sybilROCemphysema5911} in the Appendix).



However, it should be noted that the PanCan2b model, which showed significant performance disparities against lower-BMI participants, did not appear to be confounded by any of these prevalence disparities. When inspecting ROC curves, PanCan2b appeared to retain significant AUROC disparities based on BMI when isolating or excluding these confounders, as shown in Table \ref{tab:bmiIsolationPlusROCpancan}. For participants with emphysema, participants with lower BMI had an AUROC of 0.68 (0.62, 0.73), while higher-BMI participants had an AUROC of 0.80 (0.76, 0.83) ($p < 0.001$).

\subsubsection{High School Graduation Disparity}

Here, we further assessed the substantially lower specificity from all three models on participants who reported education levels lower than that of high school graduates. Examining the prevalence of risk factors between this subgroup and other participants (see Table \ref{tab:educationIsolationPlusROCpancan} in the Appendix) revealed key disparities in age, work, and smoking. Non-graduates in NLST appeared to be older than graduates (with 70\% above 61 years old, compared to 55\% of graduates), but had started smoking earlier, perhaps explaining their greater pack-years, years of smoking, and likelihood of still smoking. They were also far more likely to work in welding and with a smoker, but less likely to work without a mask. Also, 71\% of them had nodules with a diameter of 6mm or larger (compared to 62\% of graduates).

When isolating and excluding these factors, Venkadesh21 and Sybil continued to exhibit substantially more false positives for non-graduates (see Tables \ref{tab:educationTPRandTNRisolatedVenk21} and \ref{tab:educationTPRandTNRisolatedSybil} in the Appendix). However, PanCan2b's specificity disparity appeared to be confounded when splitting by age and nodule diameter, which can be seen in Table \ref{tab:educationTPRandTNRisolatedPanCan} in the Appendix. When isolating participants older than 61 years old, PanCan2b had a specificity of 0.40 (0.38, 0.42) for high school graduates and 0.33 (0.27, 0.39) for non-graduates, with slightly intersecting CIs. This was repeated for younger participants, indicating that age confounded this disparity for PanCan2b. For participants with nodules more than 6mm wide, PanCan2b achieved similar scores for graduates (specificity: 0.15 (0.14, 0.16)) and non-graduates (specificity: 0.11 (0.08, 0.15)). PanCan2b's disparity also did not persist for high school graduates and non-graduates with smaller nodules, demonstrating that PanCan2b's disparate specificity between these subgroups was confounded by nodule diameter as well as participant age.


\section{Discussion}


This study evaluated the fairness of three lung cancer risk estimation models through performance metrics, controlling for the impact of potential clinical confounders. We found four performance disparities that cannot always be explained by clinical risk factors, so we conclude that these disparities are unfair and potentially caused by biases in the algorithms or training data. First, we observed that the Sybil full-lung cancer risk model \citep{mikhael_sybil_2023} had overall lower performance for men, unrelated to prevalence disparities of potential clinical confounders. Second, the Venkadesh21 nodule risk estimation model \citep{venkadesh_deep_2021} exhibited lower sensitivity for Black Americans compared to White Americans at an operating point of 90\% specificity, which was also not found to be influenced by prevalence disparities. Third, all models appeared to have lower specificity for participants with a lower body mass index, though for Sybil, this appears confounded by the prevalence of emphysema in an LDCT. Fourth, all models exhibited more false positives for participants who had not graduated high school, though PanCan2b's disparity was confounded by nodule diameter and age. In uncovering performance biases and investigating their fairness, this analysis aims to be a step towards further research into evaluation and mitigation of problematic AI biases, as well as discussions about the fairness of diagnostic AI systems.

\subsection{Bias Between Sexes}

This study uncovered a significant performance bias in Sybil between men and women in the NLST cohort across operating points. If deployed with a clinically relevant threshold such as 90\% sensitivity or 90\% specificity, Sybil's predictions would have had higher rates of overestimation and underestimation of lung cancer risk respectively for men than for women. This result differs from other studies which found no AUROC disparities between men and women in Sybil's Year 1 score in the NLST test set from \citet{mikhael_sybil_2023} or from two American hospital screening datasets \citep{Simon2023-SybilGender}. This finding may be important since it highlights the potential for an AI model to exhibit different biases when deployed to different screening populations.

Even though Sybil was trained and evaluated on a 58\% male dataset, it appeared to have better performance for women. This is contrary to the assumptions often made in AI fairness research, where a class imbalance in data is assumed to lead to lesser performance in the underrepresented subgroup. This raised questions about the cause of these performance disparities, including whether it was related to differences in potential clinical confounders between sexes. In the NLST data, both sexes had equal rates of lung cancer diagnosis, though men tended to be taller and heavier than women, had greater rates of smoking, and were more likely to work in dangerous fields without proper respiratory protection. This reflects clinical findings in sex disparities in smoking rates \citep{Donington2011-GenderLC} and exposure to carcinogens \citep{Kreuzer2000-GenderExposureLC}. However, Sybil's sex disparity persisted in the NLST cohort when isolating for (or excluding) any of these characteristics (including BMI and weight), with the exception of height. When comparing men and women of similar height, Sybil still achieved a higher AUROC for women, but this difference was reduced and not statistically significant. The disparity in sensitivity also became less substantial, though there remained an overestimation bias for shorter men compared to shorter women. While height showed a small effect in this analysis, it is not a clinical confounder of Sybil's sex bias, because there is no strong link between height and lung cancer risk \citep{Abe2021-BMIHeightLCJapan, Le2024-heightLCnetherlands}.

Since Sybil's disparate performance for men has not been shown to be linked to potential confounders, we determine that its performance disparity is unfair, according to JustEFAB. Depending on the performance threshold applied, using Sybil's risk estimations could lead to greater false negatives or false positives for men in LCS. Sybil's bias in sex (and also height) may potentially be related to shortcut learning, since it has been shown that Sybil could accurately predict sex, height and weight from a LDCT \citep{mikhael_sybil_2023}. Conversely, the Venkadesh21 nodule-only risk estimation model performed equally for both men and women across cohorts, indicating that examining only nodules may help to prevent a bias between sexes in lung cancer risk estimation. 

This finding highlights the need for further discussion on the relationship between sex and lung cancer. While many studies have shown a link between the two, the biological connection between the two is unclear and a topic of ongoing research \citep{Donington2011-GenderLC, Gasperino2011-GenderRiskFactor, May2023-SexDifferencesLC}. As a result, some have argued that sex is an independent risk factor in LCS \citep{Nakamura2011-GenderPrognosticFactor, Gasperino2011-GenderRiskFactor, Donington2011-GenderLC}. Moreover, some studies have found greater lung cancer survival rates for women in clinical settings \citep{Nakamura2011-GenderPrognosticFactor} and LCS trials \citep{deKoning2020-NELSON, Pinsky2013-NLSTsurvival}. Sex is explicitly included in all four malignancy risk calculators in the Brock model, since it showed a significant relationship with lung cancer in the PanCan cohort \citep{mcwilliams_probability_2013}. However, it should be noted that in our study's validation set, PanCan2b yielded greater sensitivity for women, though with more false positives. This may have resulted from the coefficient attributed to sex in its logistic regression model ($\beta = 0.6011$), which systematically gives women a higher score than men when controlling for other PanCan2b factors.

\subsection{Bias Between Racial Groups}

We observed in our experiments that the Venkadesh21 model exhibited lower sensitivity with Black American participants than White American participants. We did not observe such racial disparities in PanCan2b and Sybil's performance. The Sybil Year 1 score's lack of a racial performance disparity does not fully align with AUROC scores from previous NLST analysis by \citep{mikhael_sybil_2023}, though the previous analysis may be limited by the only 5 LDCTs with a lung cancer diagnosis from Black American participants in its testing set.
When controlling for potential confounders such as hypertension and smoking status, we found that Venkadesh21's sensitivity disparity persisted. Therefore, due to a lack of explanation by potential confounders, we believe the Venkadesh21 model exhibited an unfair lower performance in Black American participants, underestimating their lung cancer risk. Considering that this model only examined nodules, we additionally investigated whether there were any differences in nodule types and size between Black and White participants. However, we did not observe any conclusive differences in nodule prevalence between these racial groups, due to the small number of samples from Black participants.

The racial bias present in this model may represent an example of health data poverty, where an underrepresented population is unable to benefit from new health technology due to insufficient data \citep{Ibrahim2021-HealthDataPoverty}. Black American participants made up only 3\% of the NLST cohort, though with a higher rate of lung cancer while also reporting lower smoking rates than White participants. This was likely a result of screening criteria which may have missed potential Black participants for NLST \citep{Aredo2022-RaceDisparityLCS}. Recent studies have shown that the median smoking rate for Black Americans with lung cancer fell below this threshold, meaning the 30-pack-year requirement for inclusion in NLST would miss the majority of lung cancer cases in this group, leading to lower sensitivity \citep{Aredo2022-RaceDisparityLCS}. Overall, a lack of sufficiently diverse data for training the Venkadesh21 model likely led it to perpetuate a lower sensitivity for Black American participants in LCS.

\subsection{Bias Relating to Body Mass Index}

In our study, the Sybil Year 1 score also demonstrated a non-significantly worse AUROC for NLST participants with a BMI below 25 $kg/m^2$ compared to those with a higher BMI, while PanCan2b performed significantly worse. However, all models had lower specificity for lower-BMI participants, with non-intersecting confidence intervals across all three clinical thresholds. This indicates lower specificity for lower-BMI participants and potentially more false positive recommendations for this group. Both in NLST and in multiple clinical studies, lower BMI was linked with higher lung cancer risk, even when controlling for smoking factors and emphysema \citep{Abe2021-BMIHeightLCJapan, Smith2012-BMIlungcancer, ElZein2013-BMIlcNoSmokingConfound}. We observed that disparities in specificity and AUROC from Sybil remain when controlling for smoking factors, indicating a lack of confounding.

However, our results showed that emphysema may have confounded Sybil's bias against lower-BMI participants, as the performance disparity reduced when isolating for participants with and without emphysema in the scan. Participants with a BMI below 25 $kg/m^2$ were more likely to have emphysema in the NLST cohort, which aligns with findings from another study which observed higher severity of emphysema for participants within this subgroup, though a causal relationship between obesity and emphysema is unknown \citep{Gu2015-BMIemphysema}. Emphysema is also a known risk factor for lung cancer and is already included in the PanCan2b model \citep{mcwilliams_probability_2013}. However, this model's disparity against lower-BMI participants did not appear to be confounded by emphysema or other factors based on prevalence disparities.

From our analysis, the included models had similar sensitivity for participants regardless of their body mass index, though with more false positives for the population more at risk for lung cancer. Since this performance disparity was linked to the higher rates of lung cancer and emphysema in lower-BMI participants, the bias may be considered less ethically significant according to JustEFAB.


\subsection{Bias Relating to High School Graduation}

Education, along with income (which is not included in NLST) often forms a key component of a person's socioeconomic status (SES), so disparate behavior against those with lower education can indicate biases based on SES. In our study, all three models demonstrated substantially lower specificity for NLST participants who had reported education levels lower than a high school diploma or a GED than participants with this or a higher education level, at a 90\% sensitivity threshold. This mirrors clinical studies which found higher false positive rates for participants with lower SES \citep{Castro2021-IncomeEdLCS}, and lower lung cancer survival rates for communities with lower rates of people with a high school diploma \citep{Erhunmwunsee2012-NeighborhoodEdLevel}. 
The models in our study did not appear to have significant AUROC disparities between these two subgroups, or substantial differences in sensitivity. 

In our NLST validation cohort, non-graduates appeared to be older than high school graduates, with longer and more intense histories of smoking, reflecting similar findings from clinical studies \citep{deGroot2018-EducationSmoking}. They were also more likely to work in welding, work with smokers, and have larger pulmonary nodules (perhaps as a result). However, the Venkadesh21 and Sybil models' specificity disparities did not appear to be confounded by any of these factors. Therefore, according to JustEFAB, these two models' higher false positive rates for non-graduates appears to be unfair.


PanCan2b's disparity was confounded by nodule diameter, which is a known risk factor included in this model, having a positive nonlinear relationship with lung cancer in PanCan2b \citep{mcwilliams_probability_2013}. Given that non-graduates in our NLST cohort are more likely to have larger nodules, this confounding appears fair according to JustEFAB. Its performance was also confounded by age, which is also factored in the model, with each additional year of a participant's age has a positive impact on the score ($\beta = 0.0287$) \citep{mcwilliams_probability_2013}. With age being a known lung cancer risk factor as well, this would appear as a fair disparity according to JustEFAB. In our NLST cohort, non-graduates tended to be older than graduates, which is surprising since they tended to start smoking at an earlier age. This may be explained by biases in the age criteria for NLST eligibility. Studies have shown these criteria to potentially have excluded younger at-risk smokers with a high school diploma or less at a higher rate than more educated younger smokers, potentially reducing both the amount of non-graduates in our cohort and affecting their correlation with age \citep{Castro2021-IncomeEdLCS}.

\subsection{Limitations}

This study is primarily limited by the narrow availability of LDCT LCS datasets that are large, demographically diverse, and contain demographic data. In the NLST, the ratio of malignant LDCT scans to benign cases is relatively low, being around 10\% in our validation datasets. While this is to be expected in a screening setting, the small number of scans with lung cancer limits our study's ability to examine smaller subgroups. This also limits our ability to examine the impact of multiple confounders on some demographic disparities (i.e. race), since splitting the NLST cohort by multiple risk factors can result in too few malignant scans within a subset for reliable analysis.

A low rate of lung cancer, combined with the fact that non-White participants only account for 7\% of the NLST dataset, prevents us from examining the models' impact on Asian Americans, Native Americans, Native Hawaiians, and Hispanic or Latino participants. This underrepresentation is tied to biases in the pack-year thresholds in screening criteria against racial minorities \citep{Aredo2022-RaceDisparityLCS} and those with a lower SES \citep{Castro2021-IncomeEdLCS}, and makes it challenging to validate whether current models are potentially unfair across diverse populations. Alternative screening criteria, such as using a risk model for determining eligibility, has shown promise in reducing this racial bias \citep{Choi2023-PLCOraceLCS}, but there is a lack of large, diverse LDCT datasets reflecting such a change. This limits the data available to examine the Venkadesh21 model's racial disparities. 

Additionally, we note that characteristics such as race, current smoking status, work history and previous diagnosis were self-reported by NLST participants and limited to the questionnaire provided by the NLST research team between 2002 and 2007. It is also important to remember that categories for race are imprecisely defined and based on the social context of where the data is collected \citep{mccraden_justefab}. While the NLST collects an extensive list of characteristics from its participant questionnaire, other screening trials may be limited by their size \citep{dlcst}, and relative lack of demographic information in the interest of privacy \citep{vanBekkum2025-AIact-GDPR}. 

\subsection{Future Work}

This study presents opportunities for multiple methods to mitigate biases in future work. For the Venkadesh21 model's racial bias, the JustEFAB framework recommends the most \textit{upstream} solution, which in this case would be collecting new screening trial data with less biased criteria and ideally a larger percentage of non-White participants \citep{mccraden_justefab}. More malignant scans from underrepresented groups can allow an intersectional analysis within racial groups, which is important since discrimination is often compounded on factors like race, sex, and socioeconomic status \citep{seyyed-kalantari_underdiagnosis_2021}. However, in the absence of new screening data, a practical mitigation strategy may be to oversample the nodules from Black American participants during training using data augmentation to reduce the data imbalance \citep{Bria2020-oversampling}.

If sex is considered to be an independent LC risk factor \citep{Gasperino2011-GenderRiskFactor}, future studies could add sex as part of Sybil's loss function, as a means to optimize performance for both men and women in training, or evaluate the utility of separate models for men and women in LCS. To reduce shortcut learning, some studies have proposed adversarial debiasing to train models to not predict certain characteristics \citep{Yang2023-AdvTraining}, though this may remove clinically relevant information useful for risk estimation \citep{Brown2023-ShortcutLearning}. There is also potential for an ensemble of models (i.e. a nodule-based model and a full-lung model), which could show promise in reducing biases and increasing overall performance \citep{Zhang2022-EnsembleMitigation}. However, it should be noted that methods to mitigate bias can flatten relevant differences across the population and may result in lower overall performance, or increase the performance for one subgroup and reduce performance for others \citep{narayanan2018translation, giovanola_beyond_2023}.

Further investigation into the explainability of AI models is also key for fairness, since it can help practitioners and participants understand why a model gives a participant a specific prediction \citep{giovanola_beyond_2023, mccraden_justefab, richardson2021framework}. This is particularly difficult in DL models, which are black-boxes due to their large numbers of parameters. Further research in developing and evaluating model explanations \citep{Nauta2023-Co12explainableAI} could uncover insights into their behavior, which may better inform radiologists into how much to trust the model's prediction for a particular scan. 

\subsection{State of Algorithmic Fairness in Screening Practice}

We as AI researchers and developers chose one particular ethical framework for this study, and evaluated disparity between subgroups on metrics used in previous LCS AI studies \citep{venkadesh_deep_2021, mikhael_sybil_2023, Simon2023-SybilGender, mcwilliams_probability_2013, Aredo2022-RaceDisparityLCS}. However, there must be careful considerations made in practice about what exactly is a fair AI tool for a screening setting. Studies interviewing AI developers have found that they are often ill-equipped to make these decisions alone \citep{richardson2021framework, Griffin2024-fr}. This is due in part to the large number of ethical definitions and frameworks for AI fairness \citep{mccraden_justefab, obermeyer2021algorithmic, giovanola_beyond_2023, barocas-hardt-narayanan}. These contain criteria which are often contradictory \citep{ferrara2023Fairness, narayanan2018translation} or too abstract and detached from technical practice \citep{richardson2021framework}. This can make fair AI development overwhelming and confusing for developers to navigate through, and often raises more questions than answers \citep{Griffin2024-fr}. 

This highlights the need for further research towards developing frameworks for fair AI decision-making in screening. Such frameworks should be backed by a shared and just definition of fairness, aligned with the expertise and values of screening stakeholders. These should also effectively translate ethical framing around AI fairness into actionable field-specific guidelines, since AI fairness and bias highly depends on the domain \citep{richardson2021framework}. In medical screening, clinicians may argue for the usage of demographic characteristics like sex or race, since they may act as a proxy for risk factors for lung cancer (i.e. environmental exposure, work history, hormone patterns) \citep{Donington2011-GenderLC, Nakamura2011-GenderPrognosticFactor, Kreuzer2000-GenderExposureLC}. Additionally, proper mechanisms should be in place in organizations for flagging AI biases that would be harmful to participants \citep{mccraden_justefab, obermeyer2021algorithmic, richardson2021framework}. Such mechanisms should encourage developers and researchers to proactively develop fair AI systems, allow them to easily report unfair biases when detected, and provide guidelines for proper bias mitigation. 
Therefore, further development of fair AI systems for LCS requires interdisciplinary research and discussions between various stakeholders to determine norms around algorithmic fairness, which are then translated into priorities and practices at the organization level.

\section{Conclusion}

We uncovered four performance disparities in lung cancer risk estimation models investigated in this work which, according to the JustEFAB ethical framework, would lead to unfair outcomes for specific participant groups. The Sybil lung-based risk estimation model performed worse for men than women, even when controlling for the prevalence of known clinical confounders, despite men making up the majority of NLST participants. The Venkadesh21 nodule risk estimation model had a lower sensitivity for Black American participants, likely due to underrepresentation in NLST. Additionally, all models found had higher false positive rates for lower-BMI participants and those without a high school diploma or GED, who are already more likely to have lung cancer than those with a higher BMI or education status. These findings raise further questions about the relationships between these demographic characteristics and lung cancer risk factors and the downstream impact of biases in lung cancer screening criteria,  and invite further research into the effectiveness of bias mitigation strategies. They also highlight the need for further discussions and interdisciplinary research on how to best apply processes to improve AI fairness in LCS, to ensure high quality diagnostic screening for all who may be at risk for lung cancer.

\acks{This study is part of the project ROBUST: Trustworthy AI-based Systems for Sustainable Growth (project number KICH3.L TP.20.006). The researchers received funding from: the Dutch Science Foundation, the Dutch Ministry of Economic Affairs, MeVis Medical Solutions (Bremen, Germany), Siemens Healthineers, the Fulbright U.S. Student Program and ELLIS. We also acknowledge important conversations with P.G.M. and E.Th.S. in this study.}


%
\ethics{The work follows appropriate ethical standards in conducting research and writing the manuscript, following all applicable laws and regulations regarding treatment of animals or human subjects.}

\coi{
    \textbf{S.G.}: no relevant conflicts of interest. 
    \textbf{M.V.}: funded by a public private project with funding from the Dutch Science Foundation, the Dutch Ministry of Economic Affairs, and MeVis Medical Solutions (Bremen, Germany).
    \textbf{A.H.}: no relevant conflicts of interest.
    \textbf{J.K.}: no relevant conflicts of interest.
    \textbf{C.J.}: research grants and royalties to host institution from MeVis Medical Solutions (Bremen, Germany); payment for lectures from Canon Medical Systems and Johnson \& Johnson; collaborator in a public-private research project where Radboud University Medical Center collaborates with Philips Medical Systems (Best, the Netherlands). 
    \textbf{L.P.}: reviewer for the FAIMI MELBA Special Fairness Issue.
    \textbf{F.v.d.G.}: funded by a public private project with funding from the Dutch Science Foundation, the Dutch Ministry of Economic Affairs, and MeVis Medical Solutions (Bremen, Germany).
}

\data{The National Lung Screening Trial (NLST) dataset is only accessible via request, as the data is not publicly available. Permission for this study was obtained from the NLST \citep{nlst} through the National Cancer Institute Cancer Data Access System (approved Project ID: NLST-1268 - Automated classification/ detection of Incidental Findings in LCS scans).}

\bibliography{refs} \label{references}


\clearpage
\appendix

\section{NLST Training Datasets for Venkadesh21 and Sybil}

While both Venkadesh21 \citep{venkadesh_deep_2021} and Sybil \citep{mikhael_sybil_2023} were trained on NLST data, they were trained on different subsets suited to the design choices made for each model. Since Venkadesh21 is a nodule risk estimation model, its authors chose a set of NLST scans with visible nodules which could be retrospectively located and annotated. The lung cancer labels for this dataset, along with our experiment's validation set, were determined by whether the patient received a lung cancer diagnosis. Venkadesh21's full NLST training dataset contains 16,077 nodules across 10,183 scans, which was used in 10-fold cross validation to create an ensemble of models. Demographic characteristics for this full set are shown in Table \ref{tab:venk21trainData}.

\begin{table*}[ht!]
\centering
\caption{Demographics of the NLST training set used for Venkadesh21 (n=10183 scans). HS = High School.}
\label{tab:venk21trainData}
\begin{tabular}{ll|rrr}
\toprule
Characteristic & Subgroup & Malignant (n=1199) & Benign (n=8984) & All Scans (n=10183) \\
\midrule
\multirow[c]{7}{*}{Education Status} & 8th grade or less & 24 (2.0) & 140 (1.6) & 164 (1.6) \\
 & 9th-11th grade & 68 (5.7) & 450 (5.0) & 518 (5.1) \\
 & Associate Degree & 295 (24.6) & 1992 (22.2) & 2287 (22.5) \\
 & Bachelors Degree & 182 (15.2) & 1403 (15.6) & 1585 (15.6) \\
 & Graduate School & 134 (11.2) & 1331 (14.8) & 1465 (14.4) \\
 & HS Graduate / GED & 300 (25.0) & 2238 (24.9) & 2538 (24.9) \\
 & Post-HS training & 171 (14.3) & 1275 (14.2) & 1446 (14.2) \\
\cline{1-5}
\multirow[c]{6}{*}{Race} & Asian & 19 (1.6) & 150 (1.7) & 169 (1.7) \\
 & Black & 56 (4.7) & 282 (3.1) & 338 (3.3) \\
 & More than one race & 10 (0.8) & 101 (1.1) & 111 (1.1) \\
 & Native American & 9 (0.8) & 34 (0.4) & 43 (0.4) \\
 & Native Hawaiian & 2 (0.2) & 31 (0.3) & 33 (0.3) \\
 & White & 1101 (91.8) & 8360 (93.1) & 9461 (92.9) \\
\cline{1-5}
\multirow[c]{2}{*}{Ethnicity} & Hispanic/Latino & 9 (0.8) & 141 (1.6) & 150 (1.5) \\
 & Not Hispanic/Latino & 1180 (98.4) & 8797 (97.9) & 9977 (98.0) \\
\cline{1-5}
\multirow[c]{2}{*}{Sex} & Female & 501 (41.8) & 3848 (42.8) & 4349 (42.7) \\
 & Male & 698 (58.2) & 5136 (57.2) & 5834 (57.3) \\
\cline{1-5}
Weight & Median (IQR) & 174 (50) & 180 (50) & 180 (50) \\
\cline{1-5}
Height & Median (IQR) & 68 (6) & 68 (6) & 68 (6) \\
\cline{1-5}
Body Mass Index & Median (IQR) & 26 (4) & 27 (6) & 27 (6) \\
\cline{1-5}
Age & Median (IQR) & 64 (8) & 62 (8) & 62 (8) \\
\cline{1-5}
\bottomrule
\end{tabular}
\end{table*}

Sybil was trained on a scan-level NLST subset, where only some scans contained visible nodules. Its labels were determined for each scan separately, with malignancy at each year based on the days between when a scan was collected and when a participant was diagnosed with lung cancer. Sybil's NLST dataset is split into training (n=28,162 scans), development (n=6,839), and test sets (n=6,282). Demographics for Sybil's training set are shown in Table \ref{tab:sybilTrainData}.

\begin{table*}[ht!]
\centering
\caption{Demographics of the NLST training set used for Sybil (n=28160 scans). HS = High School.}
\label{tab:sybilTrainData}
\begin{tabular}{ll|rrr}
\toprule
Characteristic & Subgroup & Malignant (n=1444) & Benign (n=26716) & All Scans (n=28160) \\
\midrule
\multirow[c]{7}{*}{Education Status} & 8th grade or less & 42 (2.9) & 311 (1.2) & 353 (1.3) \\
 & 9th-11th grade & 76 (5.3) & 1189 (4.5) & 1265 (4.5) \\
 & Associate Degree & 335 (23.2) & 6312 (23.6) & 6647 (23.6) \\
 & Bachelors Degree & 191 (13.2) & 4645 (17.4) & 4836 (17.2) \\
 & Graduate School & 129 (8.9) & 3847 (14.4) & 3976 (14.1) \\
 & HS Graduate / GED & 416 (28.8) & 6165 (23.1) & 6581 (23.4) \\
 & Post-HS training & 224 (15.5) & 3767 (14.1) & 3991 (14.2) \\
\cline{1-5}
\multirow[c]{6}{*}{Race} & Asian & 23 (1.6) & 552 (2.1) & 575 (2.0) \\
 & Black & 68 (4.7) & 968 (3.6) & 1036 (3.7) \\
 & More than one race & 13 (0.9) & 352 (1.3) & 365 (1.3) \\
 & Native American & 6 (0.4) & 90 (0.3) & 96 (0.3) \\
 & Native Hawaiian & 1 (0.1) & 69 (0.3) & 70 (0.2) \\
 & White & 1333 (92.3) & 24586 (92.0) & 25919 (92.0) \\
\cline{1-5}
\multirow[c]{2}{*}{Ethnicity} & Hispanic/Latino & 10 (0.7) & 505 (1.9) & 515 (1.8) \\
 & Not Hispanic/Latino & 1423 (98.5) & 26106 (97.7) & 27529 (97.8) \\
\cline{1-5}
\multirow[c]{2}{*}{Sex} & Female & 604 (41.8) & 10984 (41.1) & 11588 (41.2) \\
 & Male & 840 (58.2) & 15732 (58.9) & 16572 (58.8) \\
\cline{1-5}
Weight & Median (IQR) & 175 (50) & 180 (53) & 180 (53) \\
\cline{1-5}
Height & Median (IQR) & 68 (6) & 68 (6) & 68 (6) \\
\cline{1-5}
Body Mass Index & Median (IQR) & 26 (5) & 27 (6) & 27 (6) \\
\cline{1-5}
Age & Median (IQR) & 63 (9) & 61 (8) & 61 (8) \\
\cline{1-5}
\bottomrule
\end{tabular}
\end{table*}

Based on information provided for us in this study, we successfully recovered information about all samples from the Venkadesh21 NLST dataset, and all but three samples from the Sybil NLST dataset. We found that from the Venkadesh21 set, 4,271 LDCT scans are in Sybil's training set, 947 in its development set, and 757 in its testing set. This leaves 35,305 other scans in Sybil's datasets, and 4,208 scans from Venkadesh21's dataset not found in any of Sybil's sets. For our experiments, we used NLST validation predictions from Venkadesh21 (each prediction from only a single fold), so the dataset used for this study included the 5,912 LDCT scans in the Venkadesh21 cross-validation set which were not included in Sybil's training set. We successfully gathered predictions from all but one of these scans, resulting in 5,911 validation predictions.

Examining both Tables \ref{tab:venk21trainData} and \ref{tab:sybilTrainData} shows that the demographic characteristics are almost identical between the two models' training sets, and with the validation set used for our study, with almost all subgroups having smaller than a 2\% difference between datasets. This indicates very little population shift, and that the validation set is representative of both models' training sets. The only exception to this is that in the Sybil training set, its malignant scans contain about 4\% more scans from high school graduates and GED recipients (28.8\%) than the Venkadesh21 set (25.0\%), and about 2\% fewer malignant scans each from Bachelor's degree holders and participants who went to graduate school.

\section{Fairness Assessment Results}

Here, we report comprehensive results for assessing whether demographic performance disparities found in the subgroup performance analysis were confounded by the 10 factors with the highest prevalence disparities between subgroups. 


For Sybil's disparity between men and women, Table \ref{tab:genderIsolationPlusROC} displays the 10 factors with the greatest prevalence disparity and their impact on Sybil's AUROC for men and women. Table \ref{tab:genderTPRandTNRisolated} displays the impact of these factors on sensitivity (at 90\% specificity) and specificity (at 90\% sensitivity).


For Venkadesh21's performance disparity between White and Black participants, Table \ref{tab:raceIsolationPlusROC} displays the 10 factors with the greatest prevalence disparity and their impact on Venkdash21's AUROC for racial groups. Table \ref{tab:raceTPRandTNRisolated} displays the impact of these potential confounders on sensitivity (at 90\% specificity) and specificity (at 90\% sensitivity).

\begin{figure}[htb!]
    \centering
    \includegraphics[width=0.80\linewidth]{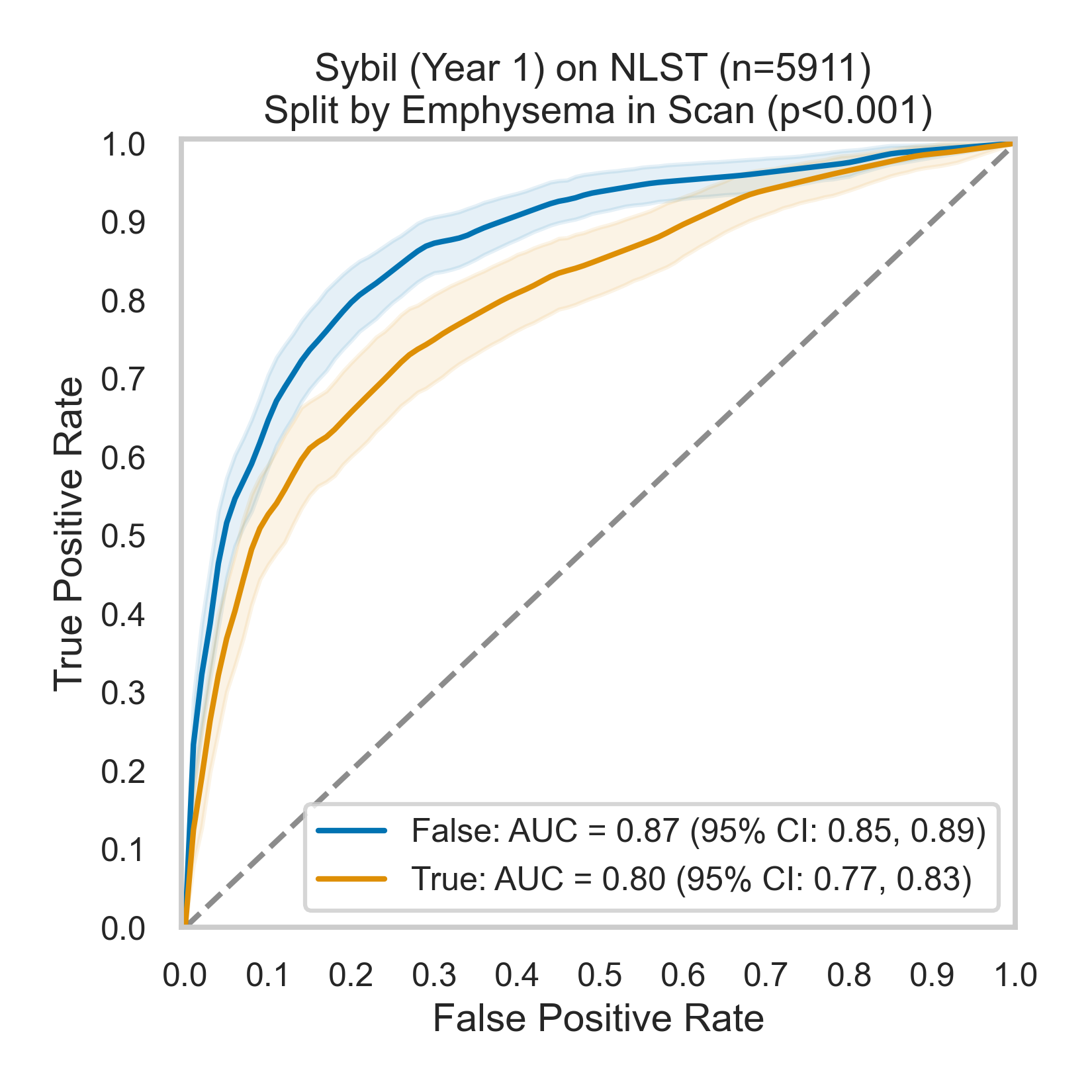}
    \caption{ROC curves (with 95\% CIs) for Sybil (Year 1), split by emphysema, in NLST (n=5911 scans).}
    \label{fig:sybilROCemphysema5911}
\end{figure}

For the disparity between participants with high BMI and those with low BMI, Table \ref{tab:bmiIsolationPlusROC} displays the 10 factors with the greatest prevalence disparity and their impact on Sybil's AUROC for these groups. Table \ref{tab:bmiTPRandTNRisolated} displays the impact of these potential confounders on sensitivity (at a 90\% specificity threshold) and specificity (at a 90\% sensitivity threshold) for Sybil.  Table \ref{tab:bmiIsolationPlusROCpancan} displays the impact of these same factors on PanCan2b's significant disparity between high-BMI and low-BMI participants. Additionally, Figure \ref{fig:sybilROCemphysema5911} displays Sybil's ROC curves reporting a significant performance disparity when emphysema is in a LDCT. 

Table \ref{tab:educationIsolationPlusROCpancan} displays the 10 factors with the highest prevalence disparities based on education status (and AUROC for PanCan2b), while sensitivity and specificity (when isolating and excluding these factors) is displayed for each model in Tables \ref{tab:educationTPRandTNRisolatedVenk21} (Venkadesh21), \ref{tab:educationTPRandTNRisolatedSybil} (Sybil) and \ref{tab:educationTPRandTNRisolatedPanCan} (PanCan2b).

\begin{sidewaystable*}[ht!]
\centering
\caption{The top 10 characteristics with the largest prevalence difference between sexes, and the AUROC scores of the Sybil (Year 1) model between men and women, isolating for them.}
\label{tab:genderIsolationPlusROC}
\begin{tabular}{ll|rrrr|rrrr|l}
\toprule
{} & {} & \multicolumn{4}{c}{Male} & \multicolumn{4}{c}{Female} & {p} \\
{Confounder} & {Subset} & {Mal} & {Ben} & {Total \%} & {AUROC} & {Mal} & {Ben} & {Total \%} & {AUROC} & {} \\
\midrule
\multirow[c]{2}{*}{Height} & $>$ 68 & 257 & 2364 & 76.2 & 0.80 (0.77, 0.83) & 10 & 136 & 5.9 & 0.84 (0.69, 0.96) & .63 \\
 & $\leqslant$ 68 & 80 & 740 & 23.8 & 0.84 (0.79, 0.88) & 234 & 2090 & 94.1 & 0.88 (0.86, 0.90) & .17 \\
\cline{1-11}
\multirow[c]{2}{*}{Weight} & $>$ 180 & 211 & 2015 & 64.7 & 0.83 (0.80, 0.86) & 36 & 554 & 23.9 & 0.86 (0.77, 0.93) & .44 \\
 & $\leqslant$ 180 & 126 & 1089 & 35.3 & 0.77 (0.72, 0.82) & 208 & 1672 & 76.1 & 0.88 (0.86, 0.90) & $<$ .001 \\
\cline{1-11}
\multirow[c]{2}{*}{Smoked Pipe} & False & 197 & 1967 & 62.9 & 0.82 (0.79, 0.85) & 239 & 2172 & 97.6 & 0.88 (0.85, 0.90) & .01 \\
 & True & 138 & 1118 & 36.5 & 0.80 (0.76, 0.84) & 5 & 40 & 1.8 & 0.97 (0.93, 1.00) & .002 \\
\cline{1-11}
\multirow[c]{2}{*}{Smoked Cigars} & True & 102 & 963 & 31.0 & 0.79 (0.74, 0.83) & 7 & 76 & 3.4 & 0.97 (0.92, 1.00) & .001 \\
 & False & 233 & 2127 & 68.6 & 0.82 (0.79, 0.86) & 237 & 2141 & 96.3 & 0.88 (0.85, 0.90) & .01 \\
\cline{1-11}
\multirow[c]{2}{*}{Work w/o Mask} & False & 228 & 2029 & 65.6 & 0.82 (0.79, 0.85) & 214 & 1936 & 87.0 & 0.89 (0.87, 0.91) & $<$ .001 \\
 & True & 109 & 1075 & 34.4 & 0.79 (0.74, 0.84) & 30 & 290 & 13.0 & 0.79 (0.69, 0.87) & .92 \\
\cline{1-11}
\multirow[c]{2}{*}{Pack-Years} & $\leqslant$ 55 & 141 & 1601 & 50.6 & 0.81 (0.77, 0.85) & 144 & 1504 & 66.7 & 0.87 (0.83, 0.90) & .06 \\
 & $>$ 55 & 196 & 1503 & 49.4 & 0.80 (0.77, 0.84) & 100 & 722 & 33.3 & 0.90 (0.86, 0.93) & $<$ .001 \\
\cline{1-11}
\multirow[c]{2}{*}{BMI} & $<$ 25 & 92 & 811 & 26.2 & 0.76 (0.70, 0.82) & 117 & 879 & 40.3 & 0.85 (0.81, 0.89) & .009 \\
 & $\geqslant$ 25 & 245 & 2293 & 73.8 & 0.83 (0.80, 0.86) & 127 & 1347 & 59.7 & 0.90 (0.87, 0.93) & .003 \\
\cline{1-11}
\multirow[c]{2}{*}{Cigarettes per Day} & $\leqslant$ 25 & 141 & 1473 & 46.9 & 0.80 (0.76, 0.84) & 139 & 1366 & 60.9 & 0.86 (0.82, 0.89) & .06 \\
 & $>$ 25 & 196 & 1631 & 53.1 & 0.82 (0.78, 0.85) & 105 & 860 & 39.1 & 0.91 (0.88, 0.94) & $<$ .001 \\
\cline{1-11}
\multirow[c]{2}{*}{Lived w/ Smoker} & False & 43 & 498 & 15.7 & 0.81 (0.74, 0.89) & 21 & 103 & 5.0 & 0.85 (0.75, 0.93) & .63 \\
 & True & 292 & 2585 & 83.6 & 0.81 (0.78, 0.84) & 222 & 2112 & 94.5 & 0.88 (0.86, 0.91) & $<$ .001 \\
\cline{1-11}
\multirow[c]{2}{*}{Pneumonia Diag.} & True & 55 & 597 & 18.9 & 0.68 (0.59, 0.77) & 81 & 649 & 29.6 & 0.86 (0.81, 0.90) & $<$ .001 \\
 & False & 281 & 2502 & 80.9 & 0.84 (0.81, 0.86) & 157 & 1574 & 70.1 & 0.90 (0.87, 0.92) & .007 \\
\cline{1-11}
\bottomrule
\end{tabular}
\end{sidewaystable*}

\begin{sidewaystable*}[ht!]
\centering
\caption{Sensitivity and Specificity at specific thresholds (with 95\% confidence intervals) for the Sybil (Year 1) model between men and women, isolating for the top 10 characteristics with the largest prevalence difference between sexes. Single asterisk (*) = metric of one subgroup is outside the CI of the other. Double asterisks (**) = CIs do not intersect.}
\label{tab:genderTPRandTNRisolated}
\begin{tabular}{ll|rrr|rrr}
\toprule
{} & {} & \multicolumn{3}{c}{Sensitivity (90\% \revision{Overall} Specificity)} & \multicolumn{3}{c}{Specificity (90\% \revision{Overall} Sensitivity)} \\
{Confounder} & {Subset} & {Male} & {Female} & {CI} & {Male} & {Female} & {CI} \\
\midrule
\multirow[c]{2}{*}{Height} & $>$ 68 & 0.52 (0.46, 0.59) & 0.60 (0.29, 0.92) &  & 0.45 (0.43, 0.47) & 0.47 (0.39, 0.56) &  \\
 & $\leqslant$ 68 & 0.55 (0.45, 0.67) & 0.66 (0.60, 0.72) &  & 0.49 (0.46, 0.53) & 0.57 (0.55, 0.59) & ** \\
\cline{1-8}
\multirow[c]{2}{*}{Weight} & $>$ 180 & 0.56 (0.49, 0.62) & 0.64 (0.47, 0.80) &  & 0.48 (0.46, 0.51) & 0.63 (0.59, 0.67) & ** \\
 & $\leqslant$ 180 & 0.48 (0.39, 0.56) & 0.66 (0.59, 0.72) & ** & 0.42 (0.39, 0.45) & 0.54 (0.52, 0.56) & ** \\
\cline{1-8}
\multirow[c]{2}{*}{Smoked Pipe} & False & 0.56 (0.49, 0.63) & 0.65 (0.59, 0.71) & * & 0.46 (0.44, 0.48) & 0.56 (0.54, 0.58) & ** \\
 & True & 0.49 (0.40, 0.57) & 1.00 (1.00, 1.00) & ** & 0.47 (0.44, 0.50) & 0.47 (0.33, 0.62) &  \\
\cline{1-8}
\multirow[c]{2}{*}{Smoked Cigars} & False & 0.56 (0.50, 0.62) & 0.65 (0.59, 0.71) & * & 0.46 (0.43, 0.48) & 0.56 (0.53, 0.58) & ** \\
 & True & 0.45 (0.36, 0.54) & 0.86 (0.50, 1.00) & * & 0.48 (0.45, 0.51) & 0.67 (0.56, 0.77) & ** \\
\cline{1-8}
\multirow[c]{2}{*}{Work w/o Mask} & False & 0.53 (0.46, 0.60) & 0.68 (0.62, 0.75) & ** & 0.48 (0.46, 0.50) & 0.57 (0.55, 0.59) & ** \\
 & True & 0.52 (0.43, 0.61) & 0.47 (0.30, 0.65) &  & 0.43 (0.40, 0.46) & 0.52 (0.46, 0.57) & ** \\
\cline{1-8}
\multirow[c]{2}{*}{Pack-Years} & $>$ 55 & 0.54 (0.47, 0.60) & 0.73 (0.64, 0.82) & ** & 0.43 (0.41, 0.46) & 0.54 (0.50, 0.58) & ** \\
 & $\leqslant$ 55 & 0.52 (0.44, 0.60) & 0.60 (0.52, 0.69) & * & 0.49 (0.47, 0.52) & 0.57 (0.55, 0.60) & ** \\
\cline{1-8}
\multirow[c]{2}{*}{BMI} & $<$ 25 & 0.50 (0.40, 0.60) & 0.66 (0.58, 0.74) & * & 0.41 (0.37, 0.44) & 0.51 (0.47, 0.54) & ** \\
 & $\geqslant$ 25 & 0.54 (0.48, 0.60) & 0.65 (0.56, 0.74) & * & 0.48 (0.46, 0.50) & 0.60 (0.57, 0.62) & ** \\
\cline{1-8}
\multirow[c]{2}{*}{Cigarettes per Day} & $>$ 25 & 0.54 (0.47, 0.61) & 0.74 (0.65, 0.82) & ** & 0.45 (0.43, 0.48) & 0.57 (0.53, 0.60) & ** \\
 & $\leqslant$ 25 & 0.51 (0.43, 0.60) & 0.59 (0.51, 0.67) &  & 0.48 (0.45, 0.50) & 0.56 (0.53, 0.58) & ** \\
\cline{1-8}
\multirow[c]{2}{*}{Lived w/ Smoker} & False & 0.51 (0.35, 0.67) & 0.57 (0.35, 0.79) &  & 0.46 (0.42, 0.50) & 0.53 (0.44, 0.63) &  \\
 & True & 0.53 (0.47, 0.59) & 0.67 (0.60, 0.73) & ** & 0.46 (0.44, 0.48) & 0.56 (0.54, 0.58) & ** \\
\cline{1-8}
\multirow[c]{2}{*}{Pneumonia Diag.} & False & 0.56 (0.50, 0.61) & 0.68 (0.61, 0.75) & * & 0.49 (0.47, 0.51) & 0.58 (0.56, 0.61) & ** \\
 & True & 0.40 (0.28, 0.53) & 0.64 (0.54, 0.75) & ** & 0.35 (0.31, 0.39) & 0.51 (0.47, 0.55) & ** \\
\cline{1-8}
\bottomrule
\end{tabular}
\end{sidewaystable*}

\begin{sidewaystable*}[ht!]
\centering
\caption{The top 10 characteristics with the largest prevalence difference between White and Black participants, and the AUROC scores of the Venkadesh21 model between these racial groups isolating for them.}
\label{tab:raceIsolationPlusROC}
\begin{tabular}{ll|rrrr|rrrr|l}
\toprule
{} & {} & \multicolumn{4}{c}{White} & \multicolumn{4}{c}{Black} & {p} \\
{Confounder} & {Subset} & {Mal} & {Ben} & {Total \%} & {AUROC} & {Mal} & {Ben} & {Total \%} & {AUROC} & {} \\
\midrule
\multirow[c]{2}{*}{Cigarettes per Day} & $\leqslant$ 25 & 246 & 2610 & 51.7 & 0.88 (0.86, 0.90) & 20 & 127 & 78.2 & 0.77 (0.67, 0.87) & .08 \\
 & $>$ 25 & 284 & 2383 & 48.3 & 0.90 (0.89, 0.92) & 8 & 33 & 21.8 & 0.94 (0.85, 1.00) & .50 \\
\cline{1-11}
\multirow[c]{2}{*}{Pack-Years} & $\leqslant$ 55 & 249 & 2877 & 56.6 & 0.88 (0.86, 0.90) & 23 & 125 & 78.7 & 0.79 (0.69, 0.88) & .10 \\
 & $>$ 55 & 281 & 2116 & 43.4 & 0.90 (0.88, 0.91) & 5 & 35 & 21.3 & 0.93 (0.83, 1.00) & .63 \\
\cline{1-11}
\multirow[c]{2}{*}{Hypertension Diag.} & True & 196 & 1617 & 32.8 & 0.87 (0.84, 0.90) & 16 & 86 & 54.3 & 0.73 (0.61, 0.85) & .04 \\
 & False & 329 & 3373 & 67.0 & 0.90 (0.89, 0.92) & 12 & 74 & 45.7 & 0.93 (0.87, 0.98) & .60 \\
\cline{1-11}
\multirow[c]{2}{*}{Current Smoker} & False & 247 & 2542 & 50.5 & 0.89 (0.87, 0.91) & 11 & 48 & 31.4 & 0.88 (0.77, 0.96) & .80 \\
 & True & 283 & 2451 & 49.5 & 0.89 (0.87, 0.91) & 17 & 112 & 68.6 & 0.76 (0.62, 0.88) & .05 \\
\cline{1-11}
\multirow[c]{2}{*}{Smoked Pipe} & True & 132 & 1127 & 22.8 & 0.90 (0.87, 0.92) & 5 & 12 & 9.0 & 0.93 (0.77, 1.00) & .75 \\
 & False & 397 & 3837 & 76.7 & 0.89 (0.87, 0.91) & 23 & 145 & 89.4 & 0.80 (0.70, 0.89) & .11 \\
\cline{1-11}
\multirow[c]{2}{*}{Sex} & Female & 218 & 2084 & 41.7 & 0.89 (0.87, 0.91) & 15 & 88 & 54.8 & 0.84 (0.75, 0.92) & .40 \\
 & Male & 312 & 2909 & 58.3 & 0.89 (0.87, 0.91) & 13 & 72 & 45.2 & 0.81 (0.66, 0.93) & .24 \\
\cline{1-11}
\multirow[c]{2}{*}{Diabetes Diag.} & True & 50 & 412 & 8.4 & 0.90 (0.86, 0.94) & 4 & 31 & 18.6 & 0.67 (0.32, 1.00) & .11 \\
 & False & 480 & 4578 & 91.6 & 0.89 (0.88, 0.90) & 24 & 129 & 81.4 & 0.83 (0.74, 0.91) & .26 \\
\cline{1-11}
\multirow[c]{2}{*}{Height} & $>$ 68 & 248 & 2389 & 47.7 & 0.89 (0.87, 0.91) & 11 & 60 & 37.8 & 0.78 (0.61, 0.93) & .19 \\
 & $\leqslant$ 68 & 282 & 2604 & 52.3 & 0.89 (0.87, 0.91) & 17 & 100 & 62.2 & 0.84 (0.76, 0.92) & .36 \\
\cline{1-11}
\multirow[c]{2}{*}{Work - Farming} & False & 485 & 4358 & 87.7 & 0.89 (0.88, 0.91) & 26 & 156 & 96.8 & 0.81 (0.71, 0.88) & .09 \\
 & True & 45 & 635 & 12.3 & 0.90 (0.85, 0.94) & 2 & 4 & 3.2 & 0.99 (1.00, 1.00) & .05 \\
\cline{1-11}
\multirow[c]{2}{*}{Age at Smoking Onset} & $\leqslant$ 16 & 320 & 2735 & 55.3 & 0.89 (0.87, 0.91) & 10 & 79 & 47.3 & 0.88 (0.80, 0.95) & .90 \\
 & $>$ 16 & 210 & 2258 & 44.7 & 0.90 (0.87, 0.92) & 18 & 81 & 52.7 & 0.78 (0.66, 0.89) & .07 \\
\cline{1-11}
\bottomrule
\end{tabular}
\end{sidewaystable*}

\begin{sidewaystable*}[ht!]
\centering
\caption{Sensitivity and Specificity at specific thresholds (with 95\% confidence intervals) for the Venkadesh21 model between between White and Black participants, isolating for the top 10 characteristics with the largest prevalence difference between racial groups. Single asterisk (*) = metric of one subgroup is outside the CI of the other. Double asterisks (**) = CIs do not intersect.}
\label{tab:raceTPRandTNRisolated}
\begin{tabular}{ll|rrr|rrr}
\toprule
{} & {} & \multicolumn{3}{c}{Sensitivity (90\% \revision{Overall} Specificity)} & \multicolumn{3}{c}{Specificity (90\% \revision{Overall} Sensitivity)} \\
{Confounder} & {Subset} &  {White} & {Black} & {CI} & {White} & {Black} & {CI} \\
\midrule
\multirow[c]{2}{*}{Cigarettes per Day} & $>$ 25 & 0.72 (0.66, 0.77) & 0.75 (0.40, 1.00) &  & 0.66 (0.64, 0.68) & 0.70 (0.53, 0.85) &  \\
 & $\leqslant$ 25 & 0.65 (0.60, 0.71) & 0.25 (0.07, 0.44) & ** & 0.66 (0.64, 0.68) & 0.74 (0.67, 0.81) & * \\
\cline{1-8}
\multirow[c]{2}{*}{Pack-Years} & $>$ 55 & 0.71 (0.65, 0.76) & 0.60 (0.00, 1.00) &  & 0.65 (0.63, 0.67) & 0.71 (0.54, 0.85) &  \\
 & $\leqslant$ 55 & 0.67 (0.61, 0.73) & 0.35 (0.16, 0.55) & ** & 0.67 (0.65, 0.69) & 0.74 (0.66, 0.82) &  \\
\cline{1-8}
\multirow[c]{2}{*}{Hypertension Diag.} & False & 0.71 (0.66, 0.76) & 0.67 (0.36, 0.92) &  & 0.67 (0.65, 0.69) & 0.72 (0.61, 0.82) &  \\
 & True & 0.65 (0.58, 0.72) & 0.19 (0.00, 0.40) & ** & 0.64 (0.62, 0.66) & 0.74 (0.65, 0.84) & * \\
\cline{1-8}
\multirow[c]{2}{*}{Current Smoker} & False & 0.67 (0.60, 0.72) & 0.36 (0.10, 0.67) & * & 0.67 (0.65, 0.68) & 0.77 (0.65, 0.89) &  \\
 & True & 0.71 (0.65, 0.76) & 0.41 (0.17, 0.67) & * & 0.65 (0.63, 0.67) & 0.71 (0.63, 0.80) &  \\
\cline{1-8}
\multirow[c]{2}{*}{Smoked Pipe} & False & 0.70 (0.65, 0.74) & 0.43 (0.23, 0.65) & ** & 0.66 (0.65, 0.68) & 0.71 (0.64, 0.78) &  \\
 & True & 0.65 (0.57, 0.72) & 0.20 (0.00, 0.67) &  & 0.65 (0.63, 0.68) & 0.92 (0.73, 1.00) & ** \\
\cline{1-8}
\multirow[c]{2}{*}{Sex} & Female & 0.72 (0.66, 0.78) & 0.40 (0.17, 0.67) & * & 0.68 (0.66, 0.70) & 0.73 (0.63, 0.82) &  \\
 & Male & 0.66 (0.61, 0.71) & 0.38 (0.12, 0.67) &  & 0.65 (0.63, 0.66) & 0.74 (0.64, 0.84) &  \\
\cline{1-8}
\multirow[c]{2}{*}{Diabetes Diag.} & False & 0.69 (0.64, 0.73) & 0.38 (0.18, 0.57) & ** & 0.66 (0.65, 0.67) & 0.73 (0.65, 0.80) &  \\
 & True & 0.72 (0.59, 0.84) & 0.50 (0.00, 1.00) &  & 0.66 (0.61, 0.70) & 0.74 (0.58, 0.88) &  \\
\cline{1-8}
\multirow[c]{2}{*}{Height} & $>$ 68 & 0.66 (0.60, 0.72) & 0.36 (0.10, 0.70) &  & 0.64 (0.62, 0.66) & 0.77 (0.65, 0.86) & * \\
 & $\leqslant$ 68 & 0.71 (0.66, 0.76) & 0.41 (0.17, 0.67) & * & 0.68 (0.66, 0.70) & 0.71 (0.62, 0.79) &  \\
\cline{1-8}
\multirow[c]{2}{*}{Work - Farming} & False & 0.69 (0.64, 0.73) & 0.38 (0.19, 0.57) & ** & 0.66 (0.65, 0.68) & 0.72 (0.65, 0.78) &  \\
 & True & 0.71 (0.56, 0.84) & 0.50 (0.00, 1.00) &  & 0.64 (0.61, 0.68) & 1.00 (1.00, 1.00) & ** \\
\cline{1-8}
\multirow[c]{2}{*}{Age at Smoking Onset} & $>$ 16 & 0.73 (0.67, 0.79) & 0.39 (0.17, 0.63) & ** & 0.67 (0.65, 0.69) & 0.75 (0.66, 0.84) &  \\
 & $\leqslant$ 16 & 0.66 (0.61, 0.72) & 0.40 (0.12, 0.71) &  & 0.65 (0.64, 0.67) & 0.71 (0.61, 0.81) &  \\
\cline{1-8}
\bottomrule
\end{tabular}
\end{sidewaystable*}

\begin{sidewaystable*}[ht!]
\centering
\caption{The top 10 characteristics with the largest prevalence difference between high and low BMI participants, and the AUROC scores of the Sybil (Year 1) model between BMI groups isolating for them.}
\label{tab:bmiIsolationPlusROC}
\begin{tabular}{ll|rrrr|rrrr|l}
\toprule
{} & {} & \multicolumn{4}{c}{$<$ 25} & \multicolumn{4}{c}{$\geqslant$ 25} & {p} \\
{Confounder} & {Subset} & {Mal} & {Ben} & {Total \%} & {AUROC} & {Mal} & {Ben} & {Total \%} & {AUROC} & {} \\
\midrule
\multirow[c]{2}{*}{Weight} & $>$ 180 & 10 & 90 & 5.3 & 0.93 (0.83, 0.99) & 237 & 2479 & 67.7 & 0.84 (0.81, 0.87) & .12 \\
 & $\leqslant$ 180 & 199 & 1600 & 94.7 & 0.81 (0.78, 0.84) & 135 & 1161 & 32.3 & 0.89 (0.86, 0.92) & .002 \\
\cline{1-11}
\multirow[c]{2}{*}{Current Smoker} & False & 74 & 642 & 37.7 & 0.82 (0.76, 0.87) & 193 & 2021 & 55.2 & 0.85 (0.82, 0.88) & .30 \\
 & True & 135 & 1048 & 62.3 & 0.81 (0.77, 0.86) & 179 & 1619 & 44.8 & 0.86 (0.84, 0.89) & .07 \\
\cline{1-11}
\multirow[c]{2}{*}{Sex} & Female & 117 & 879 & 52.4 & 0.85 (0.81, 0.89) & 127 & 1347 & 36.7 & 0.90 (0.87, 0.93) & .11 \\
 & Male & 92 & 811 & 47.6 & 0.76 (0.70, 0.82) & 245 & 2293 & 63.3 & 0.83 (0.80, 0.86) & .03 \\
\cline{1-11}
\multirow[c]{2}{*}{Emphysema in Scan} & True & 103 & 789 & 47.0 & 0.78 (0.73, 0.83) & 161 & 1103 & 31.5 & 0.81 (0.77, 0.85) & .44 \\
 & False & 106 & 901 & 53.0 & 0.84 (0.79, 0.89) & 211 & 2537 & 68.5 & 0.88 (0.86, 0.91) & .10 \\
\cline{1-11}
\multirow[c]{2}{*}{Hypertension Diag.} & True & 62 & 397 & 24.2 & 0.80 (0.73, 0.86) & 161 & 1373 & 38.2 & 0.82 (0.79, 0.85) & .55 \\
 & False & 143 & 1292 & 75.6 & 0.82 (0.78, 0.85) & 209 & 2265 & 61.7 & 0.88 (0.85, 0.90) & .02 \\
\cline{1-11}
\multirow[c]{2}{*}{Height} & $>$ 68 & 71 & 680 & 39.5 & 0.77 (0.70, 0.83) & 196 & 1820 & 50.2 & 0.82 (0.78, 0.85) & .19 \\
 & $\leqslant$ 68 & 138 & 1010 & 60.5 & 0.84 (0.81, 0.88) & 176 & 1820 & 49.8 & 0.89 (0.87, 0.92) & .04 \\
\cline{1-11}
\multirow[c]{2}{*}{Total Years of Smoking} & $>$ 40 & 142 & 983 & 59.2 & 0.79 (0.75, 0.84) & 248 & 1782 & 50.6 & 0.86 (0.84, 0.88) & .009 \\
 & $\leqslant$ 40 & 67 & 707 & 40.8 & 0.85 (0.80, 0.90) & 124 & 1858 & 49.4 & 0.84 (0.80, 0.88) & .69 \\
\cline{1-11}
\multirow[c]{2}{*}{Smoked Pipe} & False & 174 & 1403 & 83.0 & 0.83 (0.80, 0.87) & 262 & 2736 & 74.7 & 0.86 (0.84, 0.89) & .15 \\
 & True & 34 & 280 & 16.5 & 0.74 (0.63, 0.84) & 109 & 878 & 24.6 & 0.83 (0.79, 0.87) & .09 \\
\cline{1-11}
\multirow[c]{2}{*}{Diabetes Diag.} & True & 6 & 62 & 3.6 & 0.99 (1.00, 1.00) & 51 & 426 & 11.9 & 0.82 (0.75, 0.88) & $<$ .001 \\
 & False & 202 & 1625 & 96.2 & 0.81 (0.77, 0.84) & 321 & 3214 & 88.1 & 0.86 (0.84, 0.88) & .01 \\
\cline{1-11}
\multirow[c]{2}{*}{Smoked Cigars} & False & 181 & 1446 & 85.7 & 0.83 (0.80, 0.87) & 289 & 2822 & 77.5 & 0.86 (0.83, 0.89) & .23 \\
 & True & 27 & 238 & 14.0 & 0.70 (0.59, 0.81) & 82 & 801 & 22.0 & 0.83 (0.79, 0.88) & .02 \\
\cline{1-11}
\bottomrule
\end{tabular}
\end{sidewaystable*}

\begin{sidewaystable*}[ht!]
\centering
\caption{Sensitivity and Specificity at specific thresholds (with 95\% confidence intervals) for the Sybil (Year 1) model between between high and low BMI participants, isolating for the top 10 characteristics with the largest prevalence difference between subgroups. Single asterisk (*) = metric of one subgroup is outside the CI of the other. Double asterisks (**) = CIs do not intersect.}
\label{tab:bmiTPRandTNRisolated}
\begin{tabular}{ll|rrr|rrr}
\toprule
{} & {} & \multicolumn{3}{c}{Sensitivity (90\% \revision{Overall} Specificity)} & \multicolumn{3}{c}{Specificity (90\% \revision{Overall} Sensitivity)} \\
{Confounder} & {Subset} & {$\geqslant$ 25} & {$<$ 25} & {CI} & {$\geqslant$ 25} & {$<$ 25} & {CI} \\
\midrule
\multirow[c]{2}{*}{Weight} & $>$ 180 & 0.56 (0.50, 0.62) & 0.80 (0.50, 1.00) &  & 0.51 (0.49, 0.53) & 0.54 (0.44, 0.64) &  \\
 & $\leqslant$ 180 & 0.61 (0.52, 0.69) & 0.58 (0.51, 0.64) &  & 0.55 (0.52, 0.58) & 0.45 (0.43, 0.48) & ** \\
\cline{1-8}
\multirow[c]{2}{*}{Current Smoker} & False & 0.57 (0.50, 0.64) & 0.57 (0.45, 0.69) &  & 0.52 (0.50, 0.55) & 0.50 (0.46, 0.54) &  \\
 & True & 0.59 (0.52, 0.66) & 0.60 (0.52, 0.68) &  & 0.53 (0.50, 0.55) & 0.43 (0.40, 0.46) & ** \\
\cline{1-8}
\multirow[c]{2}{*}{Sex} & Female & 0.65 (0.57, 0.74) & 0.66 (0.57, 0.75) &  & 0.60 (0.57, 0.62) & 0.51 (0.47, 0.54) & ** \\
 & Male & 0.54 (0.48, 0.60) & 0.50 (0.40, 0.61) &  & 0.48 (0.46, 0.50) & 0.41 (0.37, 0.44) & ** \\
\cline{1-8}
\multirow[c]{2}{*}{Emphysema in Scan} & False & 0.60 (0.54, 0.66) & 0.60 (0.50, 0.70) &  & 0.56 (0.54, 0.58) & 0.51 (0.48, 0.55) & * \\
 & True & 0.55 (0.47, 0.63) & 0.57 (0.48, 0.67) &  & 0.44 (0.42, 0.47) & 0.40 (0.36, 0.43) & * \\
\cline{1-8}
\multirow[c]{2}{*}{Hypertension Diag.} & False & 0.63 (0.57, 0.70) & 0.56 (0.49, 0.64) &  & 0.55 (0.53, 0.57) & 0.46 (0.43, 0.49) & ** \\
 & True & 0.51 (0.43, 0.58) & 0.65 (0.53, 0.77) & * & 0.48 (0.46, 0.51) & 0.46 (0.41, 0.51) &  \\
\cline{1-8}
\multirow[c]{2}{*}{Height} & $>$ 68 & 0.54 (0.46, 0.61) & 0.49 (0.38, 0.61) &  & 0.48 (0.45, 0.50) & 0.39 (0.35, 0.43) & ** \\
 & $\leqslant$ 68 & 0.62 (0.55, 0.70) & 0.64 (0.55, 0.72) &  & 0.57 (0.55, 0.59) & 0.50 (0.48, 0.54) & ** \\
\cline{1-8}
\multirow[c]{2}{*}{Total Years of Smoking} & $>$ 40 & 0.59 (0.53, 0.65) & 0.59 (0.52, 0.67) &  & 0.49 (0.46, 0.51) & 0.42 (0.39, 0.45) & ** \\
 & $\leqslant$ 40 & 0.56 (0.47, 0.64) & 0.58 (0.47, 0.70) &  & 0.56 (0.54, 0.58) & 0.51 (0.48, 0.55) & * \\
\cline{1-8}
\multirow[c]{2}{*}{Smoked Pipe} & False & 0.60 (0.54, 0.67) & 0.61 (0.55, 0.69) &  & 0.54 (0.52, 0.56) & 0.47 (0.44, 0.49) & ** \\
 & True & 0.51 (0.42, 0.61) & 0.47 (0.30, 0.65) &  & 0.49 (0.45, 0.52) & 0.42 (0.36, 0.47) & * \\
\cline{1-8}
\multirow[c]{2}{*}{Diabetes Diag.} & False & 0.57 (0.52, 0.63) & 0.58 (0.51, 0.65) &  & 0.53 (0.51, 0.54) & 0.46 (0.43, 0.48) & ** \\
 & True & 0.61 (0.47, 0.72) & 1.00 (1.00, 1.00) & ** & 0.52 (0.48, 0.57) & 0.55 (0.42, 0.68) &  \\
\cline{1-8}
\multirow[c]{2}{*}{Smoked Cigars} & False & 0.60 (0.54, 0.66) & 0.62 (0.55, 0.68) &  & 0.53 (0.51, 0.55) & 0.46 (0.44, 0.49) & ** \\
 & True & 0.50 (0.39, 0.60) & 0.41 (0.22, 0.60) &  & 0.51 (0.48, 0.55) & 0.42 (0.36, 0.48) & * \\
\cline{1-8}
\bottomrule
\end{tabular}
\end{sidewaystable*}

\begin{sidewaystable*}[ht!]
\centering
\caption{The top 10 characteristics with the largest prevalence difference between high and low BMI participants, and the AUROC scores of the PanCan2b model between BMI groups isolating for them.}
\label{tab:bmiIsolationPlusROCpancan}
\begin{tabular}{ll|rrrr|rrrr|l}
\toprule
{} & {} & \multicolumn{4}{c}{$<$ 25} & \multicolumn{4}{c}{$\geqslant$ 25} & {p} \\
{Confounder} & {Subset} & {Mal} & {Ben} & {Total \%} & {AUROC} & {Mal} & {Ben} & {Total \%} & {AUROC} & {} \\
\midrule
\multirow[c]{2}{*}{Weight} & $>$ 180 & 10 & 90 & 5.3 & 0.85 (0.68, 0.99) & 237 & 2479 & 67.7 & 0.80 (0.77, 0.83) & .46 \\
 & $\leqslant$ 180 & 199 & 1600 & 94.7 & 0.73 (0.69, 0.76) & 135 & 1161 & 32.3 & 0.82 (0.79, 0.85) & $<$ .001 \\
\cline{1-11}
\multirow[c]{2}{*}{Current Smoker} & False & 74 & 642 & 37.7 & 0.76 (0.70, 0.82) & 193 & 2021 & 55.2 & 0.80 (0.77, 0.83) & .24 \\
 & True & 135 & 1048 & 62.3 & 0.72 (0.67, 0.77) & 179 & 1619 & 44.8 & 0.81 (0.78, 0.84) & .002 \\
\cline{1-11}
\multirow[c]{2}{*}{Sex} & Female & 117 & 879 & 52.4 & 0.74 (0.69, 0.79) & 127 & 1347 & 36.7 & 0.81 (0.78, 0.85) & .02 \\
 & Male & 92 & 811 & 47.6 & 0.73 (0.67, 0.78) & 245 & 2293 & 63.3 & 0.81 (0.79, 0.84) & .01 \\
\cline{1-11}
\multirow[c]{2}{*}{Emphysema in Scan} & True & 103 & 789 & 47.0 & 0.68 (0.62, 0.73) & 161 & 1103 & 31.5 & 0.80 (0.76, 0.83) & $<$ .001 \\
 & False & 106 & 901 & 53.0 & 0.78 (0.74, 0.83) & 211 & 2537 & 68.5 & 0.81 (0.78, 0.83) & .43 \\
\cline{1-11}
\multirow[c]{2}{*}{Hypertension Diag.} & True & 62 & 397 & 24.2 & 0.70 (0.63, 0.77) & 161 & 1373 & 38.2 & 0.78 (0.74, 0.82) & .05 \\
 & False & 143 & 1292 & 75.6 & 0.74 (0.70, 0.78) & 209 & 2265 & 61.7 & 0.82 (0.80, 0.85) & .005 \\
\cline{1-11}
\multirow[c]{2}{*}{Height} & $>$ 68 & 71 & 680 & 39.5 & 0.70 (0.64, 0.77) & 196 & 1820 & 50.2 & 0.81 (0.78, 0.84) & .004 \\
 & $\leqslant$ 68 & 138 & 1010 & 60.5 & 0.75 (0.71, 0.79) & 176 & 1820 & 49.8 & 0.81 (0.78, 0.84) & .05 \\
\cline{1-11}
\multirow[c]{2}{*}{Total Years of Smoking} & $>$ 40 & 142 & 983 & 59.2 & 0.69 (0.64, 0.74) & 248 & 1782 & 50.6 & 0.79 (0.76, 0.82) & $<$ .001 \\
 & $\leqslant$ 40 & 67 & 707 & 40.8 & 0.80 (0.74, 0.86) & 124 & 1858 & 49.4 & 0.82 (0.79, 0.85) & .56 \\
\cline{1-11}
\multirow[c]{2}{*}{Smoked Pipe} & False & 174 & 1403 & 83.0 & 0.73 (0.69, 0.77) & 262 & 2736 & 74.7 & 0.81 (0.79, 0.84) & .002 \\
 & True & 34 & 280 & 16.5 & 0.72 (0.63, 0.81) & 109 & 878 & 24.6 & 0.80 (0.76, 0.84) & .16 \\
\cline{1-11}
\multirow[c]{2}{*}{Diabetes Diag.} & True & 6 & 62 & 3.6 & 0.96 (0.91, 1.00) & 51 & 426 & 11.9 & 0.85 (0.80, 0.90) & .06 \\
 & False & 202 & 1625 & 96.2 & 0.72 (0.69, 0.76) & 321 & 3214 & 88.1 & 0.80 (0.78, 0.83) & .001 \\
\cline{1-11}
\multirow[c]{2}{*}{Smoked Cigars} & False & 181 & 1446 & 85.7 & 0.74 (0.70, 0.78) & 289 & 2822 & 77.5 & 0.80 (0.78, 0.83) & .01 \\
 & True & 27 & 238 & 14.0 & 0.66 (0.55, 0.77) & 82 & 801 & 22.0 & 0.83 (0.79, 0.87) & .008 \\
\cline{1-11}
\bottomrule
\end{tabular}
\end{sidewaystable*}

\begin{sidewaystable*}[ht!]
\centering
\caption{The top 10 characteristics with the largest prevalence difference between have graduated high school or higher and those who have not, and the AUROC scores of the PanCan2b model between these subgroups isolating for them.}
\label{tab:educationIsolationPlusROCpancan}
\begin{tabular}{ll|rrrr|rrrr|l}
\toprule
{} & {} & \multicolumn{4}{c}{$<$ HS} & \multicolumn{4}{c}{$\geqslant$ HS} & {p} \\
{Confounder} & {Subset} & {Mal} & {Ben} & {Total \%} & {AUROC} & {Mal} & {Ben} & {Total \%} & {AUROC} & {} \\
\midrule
\multirow[c]{2}{*}{Age at Smoking Onset} & $\leqslant$ 16 & 30 & 291 & 80.0 & 0.78 (0.69, 0.86) & 309 & 2580 & 53.5 & 0.77 (0.74, 0.79) & .91 \\
 & $>$ 16 & 11 & 69 & 20.0 & 0.78 (0.62, 0.92) & 217 & 2293 & 46.5 & 0.81 (0.78, 0.83) & .72 \\
\cline{1-11}
\multirow[c]{2}{*}{Total Years of Smoking} & $\leqslant$ 40 & 10 & 87 & 24.2 & 0.70 (0.49, 0.88) & 173 & 2435 & 48.3 & 0.83 (0.80, 0.85) & .15 \\
 & $>$ 40 & 31 & 273 & 75.8 & 0.81 (0.73, 0.87) & 353 & 2438 & 51.7 & 0.75 (0.73, 0.78) & .24 \\
\cline{1-11}
\multirow[c]{2}{*}{Work w/o Mask} & True & 14 & 158 & 42.9 & 0.70 (0.55, 0.85) & 123 & 1189 & 24.3 & 0.77 (0.73, 0.82) & .38 \\
 & False & 27 & 202 & 57.1 & 0.83 (0.75, 0.91) & 403 & 3684 & 75.7 & 0.79 (0.77, 0.81) & .35 \\
\cline{1-11}
\multirow[c]{2}{*}{Age} & $>$ 61 & 26 & 257 & 70.6 & 0.81 (0.71, 0.88) & 326 & 2646 & 55.0 & 0.76 (0.73, 0.78) & .36 \\
 & $\leqslant$ 61 & 15 & 103 & 29.4 & 0.77 (0.63, 0.88) & 200 & 2227 & 45.0 & 0.81 (0.78, 0.84) & .51 \\
\cline{1-11}
\multirow[c]{2}{*}{Pack-Years} & $\leqslant$ 55 & 24 & 167 & 47.6 & 0.75 (0.65, 0.84) & 254 & 2866 & 57.8 & 0.81 (0.78, 0.83) & .32 \\
 & $>$ 55 & 17 & 193 & 52.4 & 0.83 (0.70, 0.92) & 272 & 2007 & 42.2 & 0.76 (0.73, 0.79) & .30 \\
\cline{1-11}
\multirow[c]{2}{*}{Worked w/ Smoker} & True & 34 & 283 & 79.1 & 0.76 (0.67, 0.83) & 467 & 4272 & 87.8 & 0.79 (0.77, 0.81) & .52 \\
 & False & 6 & 73 & 19.7 & 0.92 (0.85, 0.97) & 57 & 561 & 11.4 & 0.76 (0.70, 0.82) & .07 \\
\cline{1-11}
\multirow[c]{2}{*}{Sex} & Female & 14 & 121 & 33.7 & 0.81 (0.68, 0.92) & 227 & 2059 & 42.3 & 0.78 (0.75, 0.81) & .70 \\
 & Male & 27 & 239 & 66.3 & 0.78 (0.69, 0.86) & 299 & 2814 & 57.7 & 0.79 (0.77, 0.82) & .79 \\
\cline{1-11}
\multirow[c]{2}{*}{Work - Welding} & False & 36 & 308 & 85.8 & 0.79 (0.71, 0.86) & 491 & 4608 & 94.4 & 0.78 (0.76, 0.80) & .91 \\
 & True & 5 & 52 & 14.2 & 0.76 (0.48, 1.00) & 35 & 265 & 5.6 & 0.83 (0.75, 0.89) & .59 \\
\cline{1-11}
\multirow[c]{2}{*}{Diameter (mm)} & $>$ 6 & 39 & 247 & 71.3 & 0.74 (0.66, 0.81) & 473 & 2917 & 62.8 & 0.73 (0.71, 0.75) & .90 \\
 & $\leqslant$ 6 & 2 & 113 & 28.7 & 0.54 (0.15, 0.91) & 53 & 1956 & 37.2 & 0.66 (0.59, 0.73) & .56 \\
\cline{1-11}
\multirow[c]{2}{*}{Current Smoker} & False & 21 & 146 & 41.6 & 0.72 (0.58, 0.83) & 238 & 2469 & 50.1 & 0.80 (0.77, 0.83) & .18 \\
 & True & 20 & 214 & 58.4 & 0.84 (0.76, 0.91) & 288 & 2404 & 49.9 & 0.77 (0.74, 0.80) & .19 \\
\cline{1-11}
\bottomrule
\end{tabular}
\end{sidewaystable*}

\begin{sidewaystable*}[ht!]
\centering
\caption{Sensitivity and Specificity at specific thresholds (with 95\% confidence intervals) for the Venkadesh21 model between between participants who have graduated high school or higher and those who have not, isolating for the top 10 characteristics with the largest prevalence difference between subgroups. Single asterisk (*) = metric of one subgroup is outside the CI of the other. Double asterisks (**) = CIs do not intersect.}
\label{tab:educationTPRandTNRisolatedVenk21}
\begin{tabular}{ll|rrr|rrr}
\toprule
{} & {} & \multicolumn{3}{c}{Sensitivity (90\% \revision{Overall} Specificity)} & \multicolumn{3}{c}{Specificity (90\% \revision{Overall} Sensitivity)} \\
{Confounder} & {Subset} & {$\geqslant$ HS} & {$<$ HS} & {CI} & {$\geqslant$ HS} & {$<$ HS} & {CI} \\
\midrule
\multirow[c]{2}{*}{Age at Smoking Onset} & $>$ 16 & 0.71 (0.65, 0.77) & 0.73 (0.44, 1.00) &  & 0.67 (0.65, 0.69) & 0.55 (0.43, 0.66) & * \\
 & $\leqslant$ 16 & 0.67 (0.61, 0.72) & 0.67 (0.50, 0.84) &  & 0.67 (0.65, 0.69) & 0.55 (0.49, 0.61) & ** \\
\cline{1-8}
\multirow[c]{2}{*}{Total Years of Smoking} & $>$ 40 & 0.70 (0.65, 0.75) & 0.71 (0.56, 0.86) &  & 0.65 (0.63, 0.67) & 0.57 (0.51, 0.63) & ** \\
 & $\leqslant$ 40 & 0.65 (0.57, 0.72) & 0.60 (0.25, 0.90) &  & 0.69 (0.67, 0.71) & 0.48 (0.38, 0.58) & ** \\
\cline{1-8}
\multirow[c]{2}{*}{Work w/o Mask} & False & 0.70 (0.65, 0.74) & 0.70 (0.51, 0.87) &  & 0.67 (0.66, 0.69) & 0.56 (0.50, 0.63) & ** \\
 & True & 0.64 (0.56, 0.72) & 0.64 (0.38, 0.90) &  & 0.68 (0.65, 0.70) & 0.53 (0.45, 0.61) & ** \\
\cline{1-8}
\multirow[c]{2}{*}{Age} & $>$ 61 & 0.71 (0.66, 0.75) & 0.69 (0.50, 0.87) &  & 0.65 (0.63, 0.66) & 0.56 (0.50, 0.62) & ** \\
 & $\leqslant$ 61 & 0.65 (0.58, 0.72) & 0.67 (0.42, 0.90) &  & 0.70 (0.68, 0.72) & 0.53 (0.44, 0.63) & ** \\
\cline{1-8}
\multirow[c]{2}{*}{Pack-Years} & $>$ 55 & 0.71 (0.66, 0.76) & 0.71 (0.44, 0.91) &  & 0.65 (0.63, 0.68) & 0.59 (0.52, 0.66) &  \\
 & $\leqslant$ 55 & 0.65 (0.59, 0.71) & 0.67 (0.48, 0.85) &  & 0.69 (0.67, 0.70) & 0.50 (0.43, 0.58) & ** \\
\cline{1-8}
\multirow[c]{2}{*}{Worked w/ Smoker} & False & 0.61 (0.49, 0.74) & 0.50 (0.00, 1.00) &  & 0.66 (0.62, 0.70) & 0.55 (0.44, 0.67) &  \\
 & True & 0.69 (0.65, 0.73) & 0.74 (0.58, 0.88) &  & 0.67 (0.66, 0.69) & 0.55 (0.49, 0.61) & ** \\
\cline{1-8}
\multirow[c]{2}{*}{Sex} & Female & 0.70 (0.64, 0.76) & 0.93 (0.78, 1.00) & ** & 0.69 (0.67, 0.71) & 0.59 (0.50, 0.67) & ** \\
 & Male & 0.68 (0.62, 0.72) & 0.56 (0.38, 0.73) &  & 0.66 (0.64, 0.68) & 0.53 (0.47, 0.59) & ** \\
\cline{1-8}
\multirow[c]{2}{*}{Work - Welding} & False & 0.68 (0.64, 0.73) & 0.67 (0.49, 0.81) &  & 0.67 (0.66, 0.69) & 0.56 (0.51, 0.62) & ** \\
 & True & 0.69 (0.54, 0.84) & 0.80 (0.33, 1.00) &  & 0.68 (0.62, 0.73) & 0.46 (0.32, 0.60) & ** \\
\cline{1-8}
\multirow[c]{2}{*}{Diameter (mm)} & $>$ 6 & 0.74 (0.70, 0.78) & 0.69 (0.53, 0.83) &  & 0.51 (0.50, 0.53) & 0.44 (0.38, 0.51) & * \\
 & $\leqslant$ 6 & 0.23 (0.12, 0.35) & 0.50 (0.00, 1.00) &  & 0.91 (0.89, 0.92) & 0.79 (0.71, 0.86) & ** \\
\cline{1-8}
\multirow[c]{2}{*}{Current Smoker} & False & 0.69 (0.63, 0.75) & 0.52 (0.31, 0.74) &  & 0.68 (0.66, 0.70) & 0.52 (0.44, 0.60) & ** \\
 & True & 0.68 (0.62, 0.74) & 0.85 (0.68, 1.00) &  & 0.66 (0.65, 0.68) & 0.57 (0.50, 0.64) & ** \\
\cline{1-8}
\bottomrule
\end{tabular}
\end{sidewaystable*}

\begin{sidewaystable*}[ht!]
\centering
\caption{Sensitivity and Specificity at specific thresholds (with 95\% confidence intervals) for the Sybil Year 1 model between between participants who have graduated high school or higher and those who have not, isolating for the top 10 characteristics with the largest prevalence difference between subgroups. Single asterisk (*) = metric of one subgroup is outside the CI of the other. Double asterisks (**) = CIs do not intersect.}
\label{tab:educationTPRandTNRisolatedSybil}
\begin{tabular}{ll|rrr|rrr}
\toprule
{} & {} & \multicolumn{3}{c}{Sensitivity (90\% \revision{Overall} Specificity)} & \multicolumn{3}{c}{Specificity (90\% \revision{Overall} Sensitivity)} \\
{Confounder} & {Subset} & {$\geqslant$ HS} & {$<$ HS} & {CI} & {$\geqslant$ HS} & {$<$ HS} & {CI} \\
\midrule
\multirow[c]{2}{*}{Age at Smoking Onset} & $>$ 16 & 0.62 (0.56, 0.68) & 0.55 (0.22, 0.83) &  & 0.52 (0.50, 0.54) & 0.39 (0.28, 0.51) & * \\
 & $\leqslant$ 16 & 0.56 (0.51, 0.62) & 0.57 (0.38, 0.74) &  & 0.51 (0.49, 0.53) & 0.36 (0.30, 0.41) & ** \\
\cline{1-8}
\multirow[c]{2}{*}{Total Years of Smoking} & $>$ 40 & 0.59 (0.53, 0.64) & 0.68 (0.52, 0.84) &  & 0.47 (0.45, 0.49) & 0.35 (0.29, 0.40) & ** \\
 & $\leqslant$ 40 & 0.59 (0.51, 0.66) & 0.20 (0.00, 0.50) & ** & 0.55 (0.54, 0.57) & 0.41 (0.31, 0.51) & ** \\
\cline{1-8}
\multirow[c]{2}{*}{Work w/o Mask} & False & 0.61 (0.56, 0.66) & 0.56 (0.38, 0.75) &  & 0.53 (0.51, 0.55) & 0.39 (0.32, 0.45) & ** \\
 & True & 0.51 (0.42, 0.60) & 0.57 (0.33, 0.83) &  & 0.46 (0.43, 0.49) & 0.34 (0.26, 0.41) & ** \\
\cline{1-8}
\multirow[c]{2}{*}{Age} & $>$ 61 & 0.61 (0.55, 0.66) & 0.65 (0.45, 0.83) &  & 0.46 (0.44, 0.48) & 0.35 (0.28, 0.41) & ** \\
 & $\leqslant$ 61 & 0.55 (0.48, 0.62) & 0.40 (0.17, 0.64) &  & 0.58 (0.55, 0.60) & 0.41 (0.32, 0.50) & ** \\
\cline{1-8}
\multirow[c]{2}{*}{Pack-Years} & $>$ 55 & 0.61 (0.55, 0.66) & 0.65 (0.38, 0.87) &  & 0.48 (0.46, 0.50) & 0.33 (0.26, 0.39) & ** \\
 & $\leqslant$ 55 & 0.57 (0.51, 0.63) & 0.50 (0.31, 0.70) &  & 0.54 (0.52, 0.55) & 0.41 (0.33, 0.48) & ** \\
\cline{1-8}
\multirow[c]{2}{*}{Worked w/ Smoker} & False & 0.49 (0.37, 0.61) & 0.50 (0.00, 1.00) &  & 0.55 (0.51, 0.59) & 0.42 (0.31, 0.54) & * \\
 & True & 0.60 (0.56, 0.64) & 0.59 (0.42, 0.74) &  & 0.51 (0.49, 0.52) & 0.35 (0.30, 0.41) & ** \\
\cline{1-8}
\multirow[c]{2}{*}{Sex} & Female & 0.64 (0.58, 0.71) & 0.79 (0.55, 1.00) &  & 0.57 (0.54, 0.59) & 0.42 (0.33, 0.50) & ** \\
 & Male & 0.55 (0.49, 0.60) & 0.44 (0.26, 0.63) &  & 0.47 (0.46, 0.49) & 0.33 (0.28, 0.40) & ** \\
\cline{1-8}
\multirow[c]{2}{*}{Work - Welding} & False & 0.59 (0.54, 0.63) & 0.56 (0.38, 0.73) &  & 0.52 (0.51, 0.53) & 0.39 (0.33, 0.44) & ** \\
 & True & 0.54 (0.38, 0.71) & 0.60 (0.00, 1.00) &  & 0.38 (0.33, 0.44) & 0.23 (0.12, 0.35) & * \\
\cline{1-8}
\multirow[c]{2}{*}{Diameter (mm)} & $>$ 6 & 0.64 (0.60, 0.68) & 0.59 (0.44, 0.74) &  & 0.38 (0.36, 0.39) & 0.26 (0.20, 0.32) & ** \\
 & $\leqslant$ 6 & 0.11 (0.03, 0.20) & 0.00 (0.00, 0.00) & ** & 0.72 (0.70, 0.74) & 0.59 (0.50, 0.68) & ** \\
\cline{1-8}
\multirow[c]{2}{*}{Current Smoker} & False & 0.59 (0.53, 0.65) & 0.43 (0.21, 0.64) &  & 0.53 (0.51, 0.55) & 0.35 (0.27, 0.43) & ** \\
 & True & 0.59 (0.53, 0.64) & 0.70 (0.48, 0.88) &  & 0.50 (0.48, 0.52) & 0.37 (0.31, 0.44) & ** \\
\cline{1-8}
\bottomrule
\end{tabular}
\end{sidewaystable*}

\begin{sidewaystable*}[ht!]
\centering
\caption{Sensitivity and Specificity at specific thresholds (with 95\% confidence intervals) for the PanCan2b model between between participants who have graduated high school or higher and those who have not, isolating for the top 10 characteristics with the largest prevalence difference between subgroups. Single asterisk (*) = metric of one subgroup is outside the CI of the other. Double asterisks (**) = CIs do not intersect.}
\label{tab:educationTPRandTNRisolatedPanCan}
\begin{tabular}{ll|rrr|rrr}
\toprule
{} & {} & \multicolumn{3}{c}{Sensitivity (90\% \revision{Overall} Specificity)} & \multicolumn{3}{c}{Specificity (90\% \revision{Overall} Sensitivity)} \\
{Confounder} & {Subset} & {$\geqslant$ HS} & {$<$ HS} & {CI} & {$\geqslant$ HS} & {$<$ HS} & {CI} \\
\midrule
\multirow[c]{2}{*}{Age at Smoking Onset} & $>$ 16 & 0.40 (0.34, 0.47) & 0.45 (0.17, 0.75) &  & 0.44 (0.42, 0.46) & 0.26 (0.16, 0.36) & ** \\
 & $\leqslant$ 16 & 0.38 (0.32, 0.43) & 0.40 (0.23, 0.59) &  & 0.46 (0.44, 0.48) & 0.38 (0.32, 0.43) & ** \\
\cline{1-8}
\multirow[c]{2}{*}{Total Years of Smoking} & $>$ 40 & 0.37 (0.32, 0.42) & 0.45 (0.28, 0.64) &  & 0.41 (0.39, 0.43) & 0.36 (0.31, 0.42) &  \\
 & $\leqslant$ 40 & 0.42 (0.35, 0.49) & 0.30 (0.00, 0.62) &  & 0.49 (0.47, 0.51) & 0.33 (0.24, 0.43) & ** \\
\cline{1-8}
\multirow[c]{2}{*}{Work w/o Mask} & False & 0.39 (0.34, 0.44) & 0.48 (0.30, 0.68) &  & 0.44 (0.43, 0.46) & 0.40 (0.34, 0.47) &  \\
 & True & 0.36 (0.28, 0.44) & 0.29 (0.08, 0.56) &  & 0.48 (0.46, 0.51) & 0.30 (0.23, 0.37) & ** \\
\cline{1-8}
\multirow[c]{2}{*}{Age} & $>$ 61 & 0.39 (0.34, 0.44) & 0.42 (0.22, 0.63) &  & 0.40 (0.38, 0.42) & 0.33 (0.27, 0.39) & * \\
 & $\leqslant$ 61 & 0.38 (0.31, 0.45) & 0.40 (0.15, 0.67) &  & 0.52 (0.50, 0.54) & 0.42 (0.33, 0.50) & * \\
\cline{1-8}
\multirow[c]{2}{*}{Pack-Years} & $>$ 55 & 0.37 (0.31, 0.43) & 0.35 (0.14, 0.60) &  & 0.45 (0.43, 0.47) & 0.38 (0.31, 0.45) & * \\
 & $\leqslant$ 55 & 0.41 (0.34, 0.47) & 0.46 (0.28, 0.67) &  & 0.46 (0.44, 0.47) & 0.33 (0.26, 0.41) & ** \\
\cline{1-8}
\multirow[c]{2}{*}{Worked w/ Smoker} & False & 0.40 (0.27, 0.54) & 0.50 (0.00, 1.00) &  & 0.43 (0.39, 0.47) & 0.45 (0.33, 0.56) &  \\
 & True & 0.39 (0.34, 0.43) & 0.41 (0.24, 0.58) &  & 0.45 (0.44, 0.47) & 0.33 (0.28, 0.39) & ** \\
\cline{1-8}
\multirow[c]{2}{*}{Sex} & Female & 0.46 (0.39, 0.52) & 0.64 (0.36, 0.88) &  & 0.37 (0.35, 0.40) & 0.26 (0.18, 0.34) & ** \\
 & Male & 0.33 (0.28, 0.39) & 0.30 (0.14, 0.47) &  & 0.51 (0.49, 0.53) & 0.41 (0.35, 0.47) & ** \\
\cline{1-8}
\multirow[c]{2}{*}{Work - Welding} & False & 0.38 (0.33, 0.42) & 0.42 (0.25, 0.57) &  & 0.45 (0.44, 0.46) & 0.36 (0.31, 0.41) & ** \\
 & True & 0.49 (0.31, 0.65) & 0.40 (0.00, 1.00) &  & 0.49 (0.43, 0.55) & 0.33 (0.20, 0.46) & * \\
\cline{1-8}
\multirow[c]{2}{*}{Diameter (mm)} & $>$ 6 & 0.43 (0.38, 0.47) & 0.44 (0.28, 0.60) &  & 0.15 (0.14, 0.16) & 0.11 (0.08, 0.15) &  \\
 & $\leqslant$ 6 & 0.00 (0.00, 0.00) & 0.00 (0.00, 0.00) &  & 0.90 (0.89, 0.92) & 0.88 (0.83, 0.94) &  \\
\cline{1-8}
\multirow[c]{2}{*}{Current Smoker} & False & 0.39 (0.33, 0.45) & 0.38 (0.19, 0.59) &  & 0.45 (0.43, 0.47) & 0.33 (0.26, 0.41) & ** \\
 & True & 0.39 (0.33, 0.44) & 0.45 (0.24, 0.68) &  & 0.45 (0.44, 0.47) & 0.37 (0.31, 0.44) & * \\
\cline{1-8}
\bottomrule
\end{tabular}
\end{sidewaystable*}




\end{document}